\documentclass[10pt,journal,compsoc]{IEEEtran}

\usepackage{eso-pic}
\usepackage{xspace}
\usepackage{times}
\usepackage{epsfig}
\usepackage{graphicx}
\usepackage{amsmath}
\usepackage{amssymb}
\usepackage{amsthm}
\usepackage{calc}
\usepackage{comment}
\usepackage{xcolor}
\usepackage{subcaption}
\usepackage{pifont}
\usepackage{colortbl}
\usepackage{setspace}
\usepackage{multirow}

\newtheorem{theorem}{Theorem}[section]

\newtheorem{lemma}[theorem]{Lemma}
\newtheorem{definition}[theorem]{Definition}

\newcommand\trainweaksup{\texttt{train-weaksup}\xspace}
\newcommand\trainfullsup{\texttt{train-fullsup}\xspace}

\newcommand\testfullsup{\texttt{test}\xspace}

\newcommand\boxacc{\texttt{BoxAcc}\xspace}
\newcommand\newboxacc{\texttt{BoxAccV2}\xspace}
\newcommand\maxboxacc{\texttt{MaxBoxAcc}\xspace}
\newcommand\newmaxboxacc{\texttt{MaxBoxAccV2}\xspace}
\newcommand\pxprec{\texttt{PxPrec}\xspace}
\newcommand\pxrec{\texttt{PxRec}\xspace}
\newcommand\pxap{\texttt{PxAP}\xspace}
\newcommand\mpxap{\texttt{mPxAP}\xspace}

\newcommand{\myparagraph}[1]{\vspace{2pt}\noindent{\bf #1}}

\newcommand\pxacc{\texttt{PxAcc}\xspace}

\newcommand\ie{\emph{i.e.}\xspace}

\newcommand\eg{\emph{e.g.}\xspace}

\newcommand\revision[1]{{\color{black} #1}}
\newcommand\addition[1]{{\color{black} #1}}
\newcommand\firstround[1]{{\color{black} #1}}

\usepackage[pagebackref=true,breaklinks=true,letterpaper=true,colorlinks,bookmarks=false]{hyperref}

\ifCLASSOPTIONcompsoc
  \usepackage[nocompress]{cite}
\else
  \usepackage{cite}
\fi

\ifCLASSINFOpdf

\else

\fi

\begin{document}

\title{Evaluation for Weakly Supervised Object Localization: Protocol, Metrics, and Datasets}

\def \kr {\textcolor{red}}
\author{Junsuk Choe\mbox{*},
	    Seong Joon Oh\mbox{*},
	    Sanghyuk Chun,
	    Seungho Lee,
	    Zeynep Akata,
        Hyunjung Shim
\IEEEcompsocitemizethanks{
\IEEEcompsocthanksitem Junsuk Choe is with the Department of Computer Science and Engineering, Sogang University, Seoul, South Korea.\protect
\IEEEcompsocthanksitem Seong Joon Oh and Sanghyuk Chun are with NAVER AI Lab.\protect
\IEEEcompsocthanksitem Seungho Lee and Hyungjung Shim are with the School of Integrated Technology, Yonsei University, South Korea.\protect
\IEEEcompsocthanksitem Zeynep Akata is with University of Tuebingen, Max Planck Institute for Informatics and Max Planck Institute for Intelligent Systems.\protect
\IEEEcompsocthanksitem Junsuk Choe and Seong Joon Oh contributed equally to this work. \protect 
\IEEEcompsocthanksitem Corresponding Authors: Hyunjung Shim \textless{}kateshim@yonsei.ac.kr\textgreater{}, and Junsuk Choe \textless{}jschoe@sogang.ac.kr\textgreater{} \protect}
}

\markboth{}
{Choe \MakeLowercase{\textit{et al.}}: Evaluation for Weakly Supervised Object Localization: Protocol, Metrics, and Datasets}

\IEEEtitleabstractindextext{%
\begin{abstract}
Weakly-supervised object localization (WSOL) has gained popularity over the last years for its promise to train localization models with only image-level labels. Since the seminal WSOL work of class activation mapping (CAM), the field has focused on how to expand the attention regions to cover objects more broadly and localize them better. However, these strategies rely on full localization supervision for validating hyperparameters and model selection, which is in principle prohibited under the WSOL setup. In this paper, we argue that WSOL task is ill-posed with only image-level labels, and propose a new evaluation protocol where full supervision is limited to only a small held-out set not overlapping with the test set. We observe that, under our protocol, the five most recent WSOL methods have not made a major improvement over the CAM baseline. Moreover, we report that existing WSOL methods have not reached the few-shot learning baseline, where the full-supervision at validation time is used for model training instead. Based on our findings, we discuss some future directions for WSOL.
Source code and dataset are available at \href{https://github.com/clovaai/wsolevaluation}{https://github.com/clovaai/wsolevaluation}.
\end{abstract}

}

\maketitle

\IEEEdisplaynontitleabstractindextext
\IEEEpeerreviewmaketitle

\section{Introduction}

As human labeling for every object is too costly and weakly-supervised object localization (WSOL) requires \textit{only} image-level labels, the WSOL research has gained significant momentum recently~\cite{CAM,ACoL,SPG,ADL,HaS,CutMix}.

\begin{figure}
	\centering
	\includegraphics[width=\linewidth]{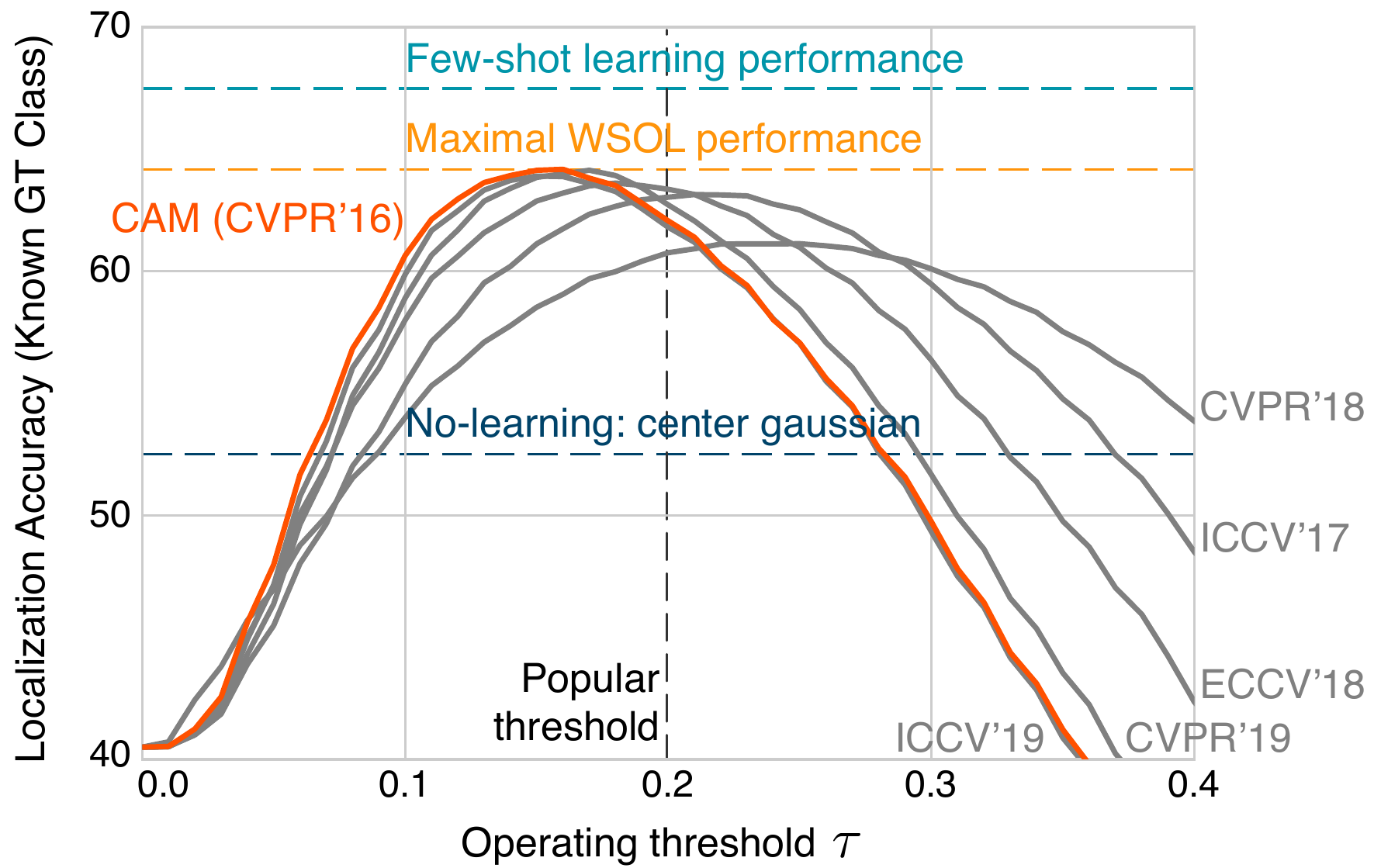}\vspace{0.3em}
	\caption{\small \textbf{WSOL 2016-2019.} Recent improvements in WSOL performances \firstround{may be an overestimation of the actual advances} due to (1) different amount of implicit full supervision through validation and (2) a fixed score-map threshold (usually $\tau=0.2$) to generate object boxes. Under our evaluation protocol with the same validation set sizes and oracle $\tau$ for each method, CAM is still the best. In fact, our few-shot learning baseline, \ie using the validation supervision (10 samples/class) at training time, outperforms existing WSOL methods. These results are obtained from ImageNet.}
	\label{fig:teaser}
	\vspace{-1.5em}
\end{figure}

Among these, class activation mapping (CAM)~\cite{CAM} uses the intermediate classifier's activations for producing score maps. The score maps represent the importance of each pixel for classification, and used for extracting bounding boxes. However, the classifier  focuses only on the most discriminative parts of the target objects. As the aim in object localization is to cover the full extent of the object, focusing only on the most discriminative parts of the objects is a limitation.
WSOL techniques since CAM have focused on this limitation and have proposed different architectural~\cite{ACoL,SPG,ADL} and data-augmentation~\cite{HaS,CutMix} solutions.
The reported state-of-the-art WSOL performances have made a significant improvement over the CAM baseline, from 49.4\% to 62.3\%~\cite{ADL} and 43.6\% to 48.7\%~\cite{ADL} top-1 localization performances on Caltech-UCSD Birds-200-2011~\cite{CUB} and ImageNet~\cite{ImageNet}, respectively.
However, these techniques have introduced a set of hyperparameters for suppressing the discriminative cues of CAM and different ways for selecting these hyperparameters.
One of such hyperparameters is the operating threshold $\tau$ for generating object bounding boxes from the score maps. Among others, the mixed policies for selecting $\tau$ has contributed to the \firstround{gradual increase in} WSOL performances over the years. \firstround{We argue that the actual qualities of score maps have not improved significantly;} see Figure~\ref{fig:teaser}.

Due to the lack of a unified definition of the WSOL task, we revisit the problem formulation of WSOL and show that WSOL problem is ill-posed in general without any localization supervision. Towards a well-posed setup, we propose a new WSOL setting where a small held-out set with full supervision is available to the learners.

Our contributions are as follows. (1) Propose new experimental protocol that uses a fixed amount of full supervision for hyperparameter search and carefully analyze six WSOL methods on three architectures and three datasets. (2) Propose new evaluation metrics as well as data, annotations, and benchmarks for the WSOL task at \href{https://github.com/clovaai/wsolevaluation}{https://github.com/clovaai/wsolevaluation}. (3) Show that WSOL has not progressed significantly since CAM, when the calibration dependency and the different amounts of full supervision are factored out. Moreover, searching hyperparameters on a held-out set consisting of 5 to 10 full localization supervision per class often leads to significantly lower performance than the few-shot learning (FSL) baselines that use the full supervision directly for model training. Finally, we suggest a shift of focus in future WSOL research: consideration of learning paradigms utilizing both weak and full supervisions, and other options for resolving the ill-posedness of WSOL (\eg background-class images).

\addition{ 
This paper is an extended version of CVPR 2020~\cite{wsoleval}. Compared to the above mentioned contributions for CVPR 2020~\cite{wsoleval}, this journal paper includes the following additional contributions:
(a) Improved metric (\newmaxboxacc) for the box-based WSOL evaluation, which considers various aspects of localization performance. 
(b) WSOL results for saliency-based explainability methods. 
(c) Analysis of classification results for WSOL methods, which shows that the classification and localization are less correlated.
(d) Further hyperparameter analysis.
(e) Few-shot learning experiments with validation. 
}

\section{Related Work}
\label{sec:rel}

\myparagraph{By model output.}
Given an input image, \textit{semantic segmentation} models generate pixel-wise class predictions~\cite{Pascal,FCN}, \textit{object detection} models~\cite{Pascal,RCNN} output a set of bounding boxes with class predictions, and \textit{instance segmentation} models~\cite{COCO,CityScapes,MaskRCNN} predict a set of disjoint masks with class \textit{and} instance labels. \textit{Object localization}~\cite{ImageNet}, on the other hand, assumes that the image contains an object of single class and produces a binary mask or a bounding box around that object coming from the class of interest. 

\myparagraph{By type of supervision.}
Since bounding box and mask labels cost significantly more than image-level labels, \eg categories~\cite{PointSup}, researchers have considered different types of localization supervision: image-level labels~\cite{papandreou2015weakly}, gaze~\cite{GazeSup}, points~\cite{PointSup}, scribbles~\cite{ScribbleSup}, boxes~\cite{BoxSup}, or a mixture of multiple types~\cite{HeterogeneousSup}. Our work is concerned with the object localization task with only image-level category labels~\cite{Oquab2015CVPR,CAM}. 

\myparagraph{By amount of supervision.}
Learning from a small amount of labeled samples per class is referred to as few-shot learning (FSL)~\cite{XCHSA19}. We recognize the relationship between our new WSOL setup and the FSL paradigm; we consider FSL methods as baselines for future WSOL methods.

\myparagraph{WSOL works.}
Class activation mapping (CAM)~\cite{CAM} turns a fully-convolutional classifier into a score map predictor by considering the activations before the global average pooling layer. Vanilla CAM has been criticized for its focus on the small discriminative part of the object.
Researchers have considered dropping regions in inputs at random~\cite{HaS, CutMix} to diversify the cues used for recognition.
Adversarial erasing techniques~\cite{ACoL,ADL} drop the most discriminative part at the current iteration.
Self-produced guidance (SPG)~\cite{SPG} is trained with auxiliary foreground-background masks generated by its own activations. \firstround{Geometry Constrained Network (GC-Net)~\cite{GCNet} trains a detector with synthesized geometric-shape masks to predict the object location directly.}
Other than object classification in static images, there exists work on localizing informative video frames for action recognition~\cite{paul2018w,liu2019completeness,xue2019danet}, but they are beyond the scope of our analysis.

\myparagraph{Relation to explainability.}
WSOL methods share similarities with the model explainability~\cite{ExplainableAI}, specifically the \textit{input attribution} task: analyzing which pixels have led to the image classification results~\cite{guidotti2019survey}. There are largely two streams of work on visual input attribution: variants of input gradients~\cite{FirstInputGradient,FirstDNNInputGradient,Wojciech16LRP,GuidedBackprop,selvaraju2017grad,IntegratedGradients,KRDCA18, park2018multimodal} and counterfactual reasoning~\cite{LIME,VedaldiMeaingfulPerturbation,zintgraf2017visualizing,ribeiro2018anchors,goyal2019counterfactual,hendricks2018grounding}. \revision{While they can be used for object localization~\cite{FirstDNNInputGradient}, they are seldom evaluated quantitatively in WSOL benchmarks. Hence, we have included them in our studies to analyze their potential as WSOL methods.}

\myparagraph{Our scope.}
We study the WSOL task, rather than weakly-supervised detection, segmentation, or instance segmentation. The terminologies tend to be mixed in the earlier works of weakly-supervised learning~\cite{WSOLasMIL2,WSOLasMIL3,OldWSOL1,WSOLasMIL4}. Extending our analysis to other weakly-supervised learning tasks is valid and will be a good contribution to the respective communities.

\begin{figure}[t]
    \centering
    \includegraphics[width=\linewidth]{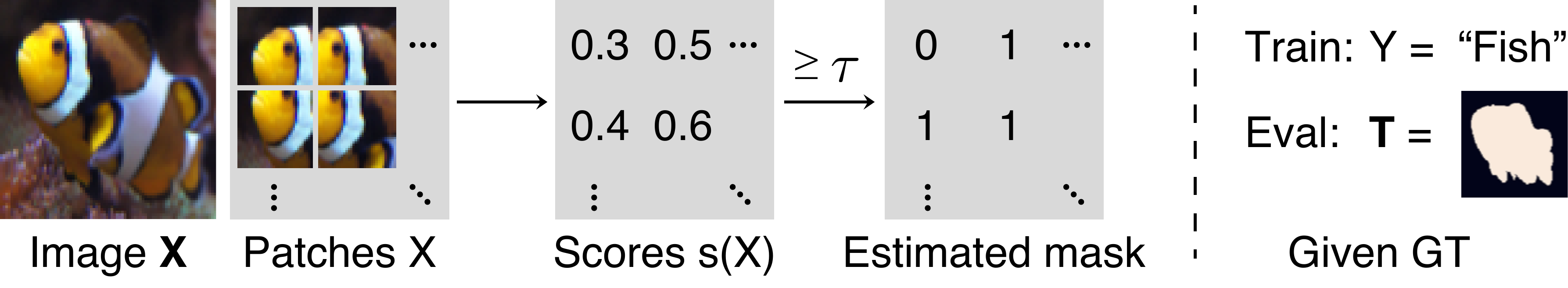}\vspace{0.3em}
    \caption{\small\textbf{WSOL as MIL.}
    WSOL is interpreted as a patch classification task trained with multiple-instance learning (MIL). The score map $s(\textbf{X})$ is thresholded at $\tau$ to estimate the mask $\mathbf{T}$.}
    \label{fig:overview}
    \vspace{-1.5em}
\end{figure}

\section{Problem Formulation of WSOL}
\label{sec:wsol_impossibility}

We define and formulate the weakly-supervised object localization (WSOL) task as an image patch classification and show the ill-posedness of the problem. We will discuss possible modifications to resolve the ill-posedness in theory.

\subsection{WSOL task as multiple instance learning}
\label{subsec:what_is_wsol}

Given an image $\mathbf{X}\in\mathbb{R}^{H\times W}$, \textbf{object localization} is the task to identify whether or not the pixel belongs to the object of interest, represented via dense binary mask $\mathbf{T}=(T_{11},\cdots,T_{HW})$ where $T_{ij}\in\{0,1\}$ and $(i,j)$ indicate the pixel indices. When the training set consists of precise image-mask pairs $(\mathbf{X},\mathbf{T})$, we refer to the task as \textbf{fully-supervised object localization (FSOL)}. In this paper, we consider the case when only an image-level label $Y\in\{0,1\}$ for global presence of the object of interest is provided per training image $\mathbf{X}$. This task is referred to as the \textbf{weakly-supervised object localization (WSOL)}.

One can treat an input image $\mathbf{X}$ as a bag of stride-1 sliding window patches of suitable side lengths, $h$ and $w$: $(X_{11},\cdots,X_{HW})$ with $X_{ij}\in\mathbb{R}^{h\times w}$. The object localization task is then the problem of predicting the object presence $T_{ij}$ at the image patch $X_{ij}$. The weak supervision imposes the requirement that each training image $\mathbf{X}$, represented as $(X_{11},\cdots,X_{HW})$, is only collectively labeled with a single label $Y\in\{0,1\}$ indicating whether at least one of the patches represents the object. This formulation is an example of the multiple-instance learning (MIL)~\cite{MIL}, as observed by many traditional WSOL works~\cite{papandreou2015weakly,WSOLasMIL2,WSOLasMIL3,WSOLasMIL4}.

Following the patch classification point of view, we formulate WSOL task as a mapping from patches $X$ to the binary labels $T$ (indices dropped). We assume that the patches $X$, image-level labels $Y$, and  the pixel-wise labeling $T$ in our data arise in an i.i.d. fashion from the joint distribution $p(X,Y,T)$. See Figure~\ref{fig:overview} for an overview. The aim of WSOL is to produce a scoring function $s(X)$ such that thresholding it at $\tau$ closely approximates the binary label $T$.

\addition{
\subsection{Case study: CAM as MIL}

In light of the above discussion, we re-interpret CAM~\cite{CAM} and its variants, a representative class of methods for WSOL, as a patch classifier trained under the MIL objective.

CAM~\cite{CAM} use the scoring rules based on the posterior $s(X)=p(Y|X)$.
Originally, CAM is a technique applied on a convolutional neural network classifier $h:\mathbb{R}^{3\times H \times W}\rightarrow\mathbb{R}^{C}$, where $C$ is the number of classes, of the following form:
\begin{equation}
h_{c}(X)=\sum_{d}W_{cd}\left(\frac{1}{HW}\sum_{ij}g_{dij}(\mathbf{X})\right)
\end{equation}
where $c,d$ are the channel-dimension indices and $i,j$ are spatial-dimension indices. In other words, $h$ is a fully convolutional neural network, followed by a global average pooling (GAP) and a linear (fully-connected) layer into a $C$-dimensional vector. We may swap the GAP and linear layers without changing the representation:
\begin{align}
h_{c}(X)
&=\frac{1}{HW}\sum_{ij}\left(\sum_{d}W_{cd}g_{dij}(\mathbf{X})\right) \\
&=:\frac{1}{HW}\sum_{ij}f_{cij}(\mathbf{X}) 
\end{align}
where $f$ is now a fully-convolutional network. Each pixel $(i,j)$ in the feature map, $\left(f_{1ij}(\mathbf{X}),\cdots,f_{Cij}(\mathbf{X})\right)$, corresponds to the classification result of the corresponding field of view in the input $\mathbf{X}$, written as $X_{ij}$. Thus, we equivalently write \begin{align}
h_{c}(X)=\frac{1}{HW}\sum_{ij}f_{c}(X_{ij}) 
\end{align}
where $f$ is now re-defined as a image patch classifier with 1-dimensional feature output (not fully convolutional).

The bag of patch-wise classification scores $f_{c}(X_{ij})$ is then supervised by the error between the mean outputs $h_{c}(X)$ and the ground truth label $Y$, measured by the softmax cross-entropy loss:
\begin{equation}
    \log p(Y|\mathbf{X}):=
    \log \text{softmax}^Y\left(\frac{1}{HW}\sum_{ij}f(X_{ij})\right).
\end{equation}
In other words, CAM trains the network for patch-wise scores $f_{c}(X_{ij})$ to estimate the image-wide posterior $p(Y|\mathbf{X})$. 

At inference time, CAM estimates the pixel-wise posterior $p(Y|X_{ij})$ approximately by performing a score normalization for $f_Y(X_{ij})$ (Table~\ref{tab:prior_wsol_operating_thresholds}).
}

\subsection{When is WSOL ill-posed?}
\label{subsec:when_is_wsol_unsolvable}

We show that if background cues are more strongly associated with the target labels $T$ than some foreground cues, the localization task cannot be solved, even when we know the exact posterior $p(Y|X)$ for the image-level label $Y$. We will make some strong assumptions in favor of the learner, and then show that WSOL still cannot be perfectly solved. 

We assume that there exists a finite set of \textbf{cue} labels $\mathcal{M}$ containing all patch-level concepts in natural images. For example, patches from a duck image are one of \{duck's head, duck's feet, sky, water, $\cdots$\} (see Figure~\ref{fig:duck_feet_lake}). We further assume that every patch $X$ is equivalently represented by its cue label $M(X)\in\mathcal{M}$. Therefore, from now on, we write $M$ instead of $X$ in equations and examine the association arising in the joint distribution $p(M,Y,T)$. 
We write $M^{\text{fg}},M^{\text{bg}}\in\mathcal{M}$ for foreground and background cues.

\newcommand{\eqd}{\text{duck}}
\newcommand{\eqnd}{\text{duck}^c}
\newcommand{\eqf}{\text{feet}}
\newcommand{\eql}{\text{water}}

\begin{figure}[t]
    \centering
    \includegraphics[width=1\linewidth]{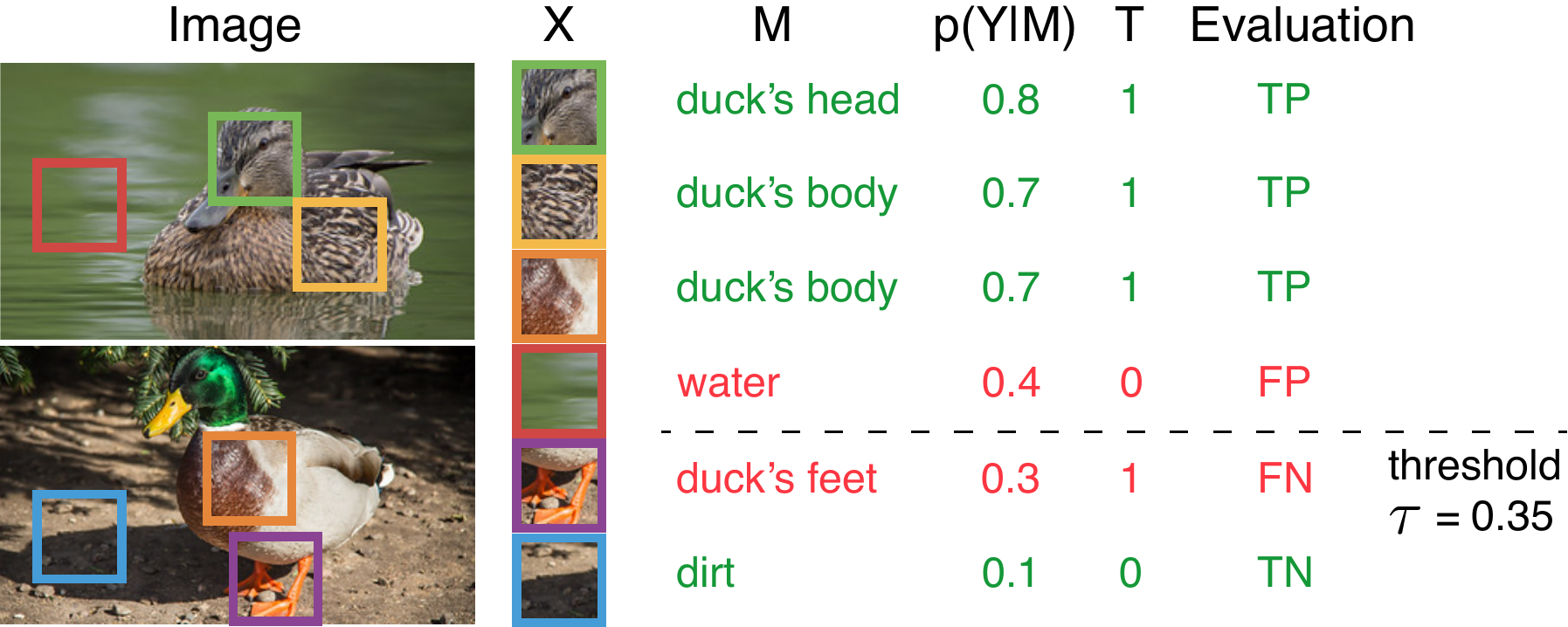}\vspace{0.3em}
    \caption{\small\textbf{Ill-posed WSOL: An example.} Even the true posterior $s(M)=p(Y|M)$ may not lead to the correct prediction of $T$ if background cues are more associated with the class than the foreground cues (\eg $p(\eqd|\eql)> p(\eqd|\eqf)$).}
    \label{fig:duck_feet_lake}
    \vspace{-1.5em}
\end{figure}

\addition{We first define an evaluation metric for our score map for an easier argumentation. 
\begin{definition}
For a scoring rule $s$ and a threshold $\tau$, we define the \textbf{pixel-wise localization accuracy} $\text{\pxacc}(s,\tau)$ as the probability of correctly predicting the pixel-wise labels:
\begin{align}
    \text{\pxacc}(s,\tau)=
    {P}_{X,T}(s(X)\geq\tau\mid T=1)\cdot{P}_{X,T}(T=1)
    \nonumber\\
    +{P}_{X,T}(s(X)<\tau\mid T=0)\cdot{P}_{X,T}(T=0)
    \nonumber
\end{align}
\end{definition}
}

We argue that, even with access to the joint distribution $p(Y,M)$, it may not be possible to make perfect predictions for the patch-wise labels $T(M)$.
\begin{lemma}
    Assume that the true posterior $p(Y|M)$ with a continuous pdf is used as the scoring rule $s(M)=p(Y|M)$. Then, there exists a scalar $\tau\in\mathbb{R}$ such that $s(M)\geq\tau$ is identical to $T$ if and only if the foreground-background posterior ratio $\frac{p(Y=1|M^{\text{fg}})}{p(Y=1| M^{\text{bg}})}\geq 1$ almost surely, conditionally on the event $\{T(M^{\text{fg}})=1\text{ and }T(M^{\text{bg}})=0\}$.
\end{lemma}
\addition{\begin{proof}
    We write $E:=\{T(M^{\text{fg}})=1\text{ and }T(M^{\text{bg}})=0\}$.
    
    \noindent
    (\textbf{Proof for ``if''}) Assume the posterior ratio $\alpha\geq 1$ almost surely, given $E$. Let
    \begin{align}
        \tau:=\min_{G:P(G\Delta \{T(m)=0\})=0}\,\,
        \max_{m\in G}\,\,
        p(Y=1| M=m)
    \end{align} 
    where $\Delta$ is the set XOR operation: $A\Delta B:=(A\cup B)\setminus (A\cap B)$. Then, for almost all $M^{\text{fg}},M^{\text{bg}}$ following $E$, 
    \begin{align}
        p(Y=1| M^{\text{fg}})\geq \tau \geq p(Y=1| M^{\text{bg}}).
        \label{eq:co_occurrence_ordered}
    \end{align}
    Therefore, 
    \begin{align}
        &P(p(Y=1| M^{\text{fg}})\geq \tau\mid T(M^{\text{fg}})=1)\nonumber\\
        &=P(p(Y=1| M^{\text{bg}})\leq \tau\mid T(M^{\text{bg}})=0) =1
        \label{eq:loc_conditionals_prob_1}
    \end{align}
    and so $\text{\pxacc}(p(Y|M),\tau)=1$.
    
    \noindent
    (\textbf{Proof for ``only if''}) Assume $\text{\pxacc}(p(Y| M),\tau)=1$ for some $\tau$. W.L.O.G., we assume that ${P}(T(M)=1)\neq 0$ and ${P}(T(M)=0)\neq 0$ (otherwise, ${P}(E)=0$ and the statement is vacuously true). Then, Equation~\ref{eq:loc_conditionals_prob_1} must hold to ensure $\text{\pxacc}(p(Y| M),\tau)=1$. Equation~\ref{eq:co_occurrence_ordered} then also holds almost surely, implying $\alpha\geq 1$ almost surely.
\end{proof}}

\begin{figure}[t]
    \centering
    \includegraphics[width=\linewidth]{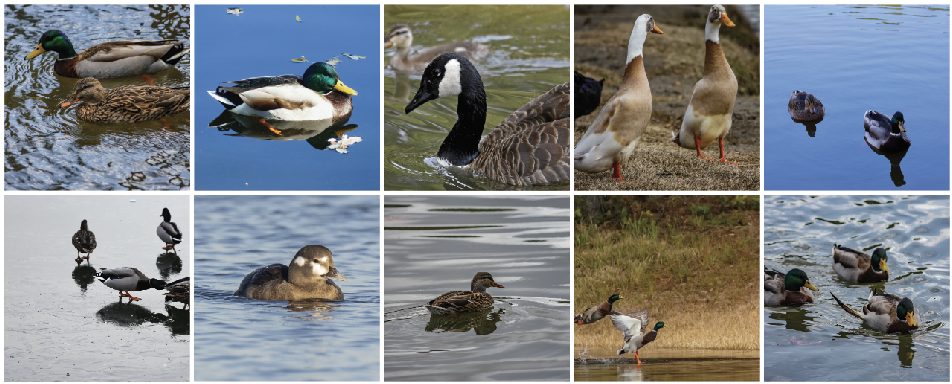}
    \caption{\small\textbf{Ducks.} Random duck images on Flickr. They contain more lake than feet pixels: $p(\eql|\eqd)\gg p(\eqf|\eqd)$.}
    \label{fig:duck_feet_lake_appendix}
\end{figure}

In other words, if the posterior likelihood for the image-level label $Y$ given a foreground cue $M^{\text{fg}}$ is less than the posterior likelihood given background $M^{\text{bg}}$ for some foreground and background cues, no WSOL method can make a correct prediction. This pathological scenario is described in Figure~\ref{fig:duck_feet_lake}: Duck's feet are less seen in duck images than the water background. Such cases are abundant in user-collected data (Figure~\ref{fig:duck_feet_lake_appendix}).

\addition{\myparagraph{Foreground-background posterior ratio}
We have described the the pathological scenario for WSOL as when the foreground-background posterior ratio $\alpha$ is small. We discuss in greater detail what it means and whether there are data-centric approaches to resolve the issue. For quick understanding, assume the task is the localization of duck pixels in images. The foreground cue of interest $M^{\text{fg}}$ is ``feet'' of a duck and background cue of interest $M^{\text{bg}}$ is ``water''. Then, we can write the posterior ratio as

\begin{equation}
\alpha:=
\frac{p(\eqd|\eqf)}{p(\eqd|\eql)}= 
\frac{p(\eqf|\eqd)}{p(\eql|\eqd)} \cdot \left(\frac{p(\eqf)}{p(\eql)}\right)^{-1}
\nonumber
\end{equation}
$\alpha<1$ implies that lake patches are more abundant in duck images than are duck's feet (see Figure~\ref{fig:duck_feet_lake_appendix}) for an illustration.

To increase $\alpha$, two approaches can be taken. (1) Increase the likelihood ratio $\frac{p(\eqf|\eqd)}{p(\eql|\eqd)}$. This can be done by collecting more images where duck's feet have more pixels than lake does. (2) Decrease the prior ratio $\frac{p(\eqf)}{p(\eql)}$. Note that the prior ratio can be written
\begin{equation}
\footnotesize
\frac{p(\eqf)}{p(\eql)}=
\frac{p(\eqf|\eqd)p(\eqd)+p(\eqf|\eqnd)p(\eqnd)}%
{p(\eql|\eqd)p(\eqd)+p(\eql|\eqnd)p(\eqnd)}
\nonumber
\end{equation}
With fixed likelihoods $p(\eqf|\eqd)$ and $p(\eql|\eqd)$, one can decrease the prior ratio by increasing the likelihood of lake cues in non-duck images $p(\eql|\eqnd)$. We can alter WSOL into a more well-posed task also by including many background images containing confusing background cues.

These data-centric approaches are promising future research directions for turning WSOL into a well-posed task.}

\myparagraph{How have WSOL methods addressed the ill-posedness?}
Previous solutions for WSOL have sought architectural modifications~\cite{ACoL, SPG, ADL} and data augmentation~\cite{HaS,CutMix} schemes that typically require heavy hyperparameter search and model selection, which are a form of implicit full supervision. For example, \cite{HaS} has found the operating threshold $\tau$ via ``observing a few qualitative results'', while others have evaluated their models over the test set to select reasonable hyperparameter values (Table 1 of \cite{HaS}, Table 6 of \cite{ACoL}, and Table 1 of \cite{ADL}). \cite{SPG} has performed a ``grid search'' over possible values. We argue that certain level of localization labels are inevitable for WSOL. In the next section, we propose to allow a fixed number of fully labeled samples for hyperparameter search and model selection for a more realistic evaluation.

\section{Evaluation Protocol for WSOL}
\label{sec:evaluation}

We reformulate the WSOL evaluation based on our observation of the ill-posedness. We define performance metrics, benchmarks, and the hyperparameter search procedure.

\subsection{Evaluation metrics}
\label{subsec:evaluation_metrics}

The aim of WSOL is to produce score maps, where their pixel value $s_{ij}$ is higher on foreground $T_{ij}=1$ and lower on background $T_{ij}=0$ (\S\ref{subsec:what_is_wsol}). We discuss how to quantify the above conditions and how prior evaluation metrics have failed to clearly measure the relevant performance. We then propose the \maxboxacc and \pxap metrics for bounding box and mask ground truths, respectively.

The \textit{localization accuracy}~\cite{ImageNet} metric entangles classification and localization performances by counting the number of images where both tasks are performed correctly. We advocate the measurement of localization performance alone, as the goal of WSOL is to localize objects (\S\ref{subsec:what_is_wsol}) and not to classify images correctly. To this end, we only consider the score maps $s_{ij}$ corresponding to the ground-truth classes in our analysis. Metrics based on such are commonly referred to as the \textit{GT-known} metrics~\cite{HaS,ACoL,SPG,ADL}.

\begin{table}[t]
	\newcommand\maxnormf{\overline{s}_{ij}}
	\newcommand\minmaxnormf{\widehat{s}_{ij}}
	\small
	\centering
	\addition{\begin{tabular}{*{2}{l}*{2}{l}}
		Method && Paper & Code  \\
		\cline{1-1} \cline{3-4}
		\vspace{-1em} & \\
		\cline{1-1} \cline{3-4}
		\vspace{-1em} & \\
		CAM~\cite{CAM} && $\maxnormf\geq 0.2$ & $\maxnormf\geq 0.2$ \\
		HaS~\cite{HaS} && Follow CAM$^\dagger$ & Follow CAM \\
		ACoL~\cite{ACoL} && Follow CAM & $\minmaxnormf\geq \text{unknown}$ \\
		SPG~\cite{SPG} && Grid search threshold & $\minmaxnormf\geq \text{unknown}$ \\
		ADL~\cite{ADL} && Not discussed & $\minmaxnormf\geq 0.2^\dagger$ \\
		CutMix~\cite{CutMix} && $\maxnormf\geq 0.15$ & $\minmaxnormf\geq 0.15$ \\
		\cline{1-1} \cline{3-4}
		\vspace{-1em} & \\
		Our protocol && $\minmaxnormf\geq \tau^\star$ & $\minmaxnormf\geq \tau^\star$ \\
		\cline{1-1} \cline{3-4}
	\end{tabular}
	\[
	\overline{s}_{ij}:=\frac{s_{ij}}{\max_{kl}s_{kl}}\quad \quad \widehat{s}_{ij}:=\frac{s_{ij}-\min_{kl}s_{kl}}{\max_{kl}s_{kl}-\min_{kl}s_{kl}}
	\]}
	\caption{\small \addition{\textbf{Calibration and thresholding in WSOL.} Score calibration is done per image: max ($\overline{s}_{ij}$) or min-max ($\widehat{s}_{ij}$) normalization. Thresholding is required only for the box evaluation. $\tau^\star$ is the optimal threshold (\S\ref{subsec:evaluation_metrics} in main paper). Daggers ($^\dagger$) imply that the threshold depends on the backbone architecture.}}
	\label{tab:prior_wsol_operating_thresholds}
\end{table}

A common practice in WSOL is to normalize the score maps per image because the maximal (and minimal) scores differ vastly across images. \revision{Prior WSOL papers have introduced either max normalization (dividing through by $\max_{ij}s_{ij}$) or min-max normalization (additionally mapping $\min_{ij}s_{ij}$ to zero). We summarize how prior work calibrates and thresholds the score maps in Table~\ref{tab:prior_wsol_operating_thresholds}. In this paper, we \firstround{mainly} use the min-max normalization.}

After normalization, WSOL methods threshold the score map at $\tau$ to generate a tight box around the binary mask $\{(i,j)\mid s_{ij}\geq \tau\}$. WSOL metrics then measure the quality of the boxes. $\tau$ is typically treated as a fixed value~\cite{CAM,ACoL,CutMix} or a hyperparameter to be tuned~\cite{HaS,SPG,ADL}.
We argue that the former is misleading because the ideal threshold $\tau$ depends heavily on the data and model architecture and fixing its value may be disadvantageous for certain methods. To fix the issue, we propose new evaluation metrics that are independent of the threshold $\tau$.

\myparagraph{Masks: \firstround{\mpxap}.}
When masks are available for evaluation, we measure the pixel-wise precision and recall~\cite{PixelPrecisionRecall}. Unlike single-number measures like mask-wise IoU, those metrics allow users to choose the preferred operating threshold $\tau$ that provides the best precision-recall trade-off for their downstream applications. We define the \textbf{pixel precision and recall at threshold} $\tau$ \textbf{and class} $c$ as:
\vspace{-.1em}
{\small
\begin{align}
    \text{\pxprec}(\tau)=
    \frac{|\{s^{(n)}_{ij}\geq\tau\}\cap\{T^{(n)}_{ij}=1\}|}%
    {|\{s^{(n)}_{ij}\geq\tau\}|}\\
    \text{\pxrec}(\tau)=
    \frac{|\{s^{(n)}_{ij}\geq\tau\}\cap\{T^{(n)}_{ij}=1\}|}%
    {|\{T^{(n)}_{ij}=1\}|}
\end{align}
\vspace{-.1em}
}%
For threshold independence, we define and use the \textbf{pixel average precision}, $\text{\pxap}:=\sum_l \text{\pxprec}(\tau_l)(\text{\pxrec}(\tau_l)-\text{\pxrec}(\tau_{l-1}))$, the area under curve of the pixel precision-recall curve. \firstround{Finally, to ensure that all the classes contribute equally to the final performance, we define \textbf{mean pixel average precision} (\mpxap) by taking the mean of the class-wise \pxap values. We use the \mpxap as the final metric in this paper.}

\myparagraph{Bounding boxes: \newmaxboxacc.}
Pixel-wise masks are expensive to collect; many datasets only provide box annotations. Since it is not possible to measure exact pixel-wise precision and recall with bounding boxes, we suggest a surrogate in this case. Given the ground truth box $B$, we define the \textbf{box accuracy at score map threshold} $\tau$ \textbf{and IoU threshold} $\delta$, \newboxacc$(\tau, \delta)$~\cite{CAM,ImageNet}, as:
{\small
\begin{align}
    \text{\newboxacc}(\tau, \delta)=
    \frac{1}{N}
    \sum_n
    1_{
    \text{IoU}\left(
    \text{boxes}(s(\mathbf{X}^{(n)}),\tau),B^{(n)}
    \right)\geq \delta
    }
\end{align}
}%
where $\text{boxes}(s(\mathbf{X}^{(n)}),\tau)$ is \revision{the set of tightest boxes around each connected component} of the mask $\{(i,j)\mid s(X^{(n)}_{ij})\geq\tau\}$. \revision{$\text{IoU}\left(\text{boxes}_A,\text{boxes}_B\right)$ is defined as the best (maximal) value among the IoUs across the sets $\text{boxes}_A$ and $\text{boxes}_B$.} For score map threshold independence, we report the box accuracy at the optimal threshold $\tau$, the \textbf{maximal box accuracy} $\text{\newmaxboxacc}(\delta):=\max_\tau\text{\newboxacc}(\tau, \delta)$, as the final performance metric. \revision{We average the performance across $\delta\in\{0.3, 0.5, 0.7\}$ to address diverse demands for localization granularity.}

\revision{\myparagraph{Comparison with the previous \maxboxacc.} The previous version presented in the conference paper~\cite{wsoleval} is deprecated. \newmaxboxacc is better in two aspects. 
(1) \maxboxacc measures the performance at a fixed IoU threshold ($\delta=0.5$), only considering a specific level of granularity for localization outputs.
(2) \maxboxacc takes the \textit{largest} connected component for estimating the box, assuming that the object of interest is usually large. \newmaxboxacc removes this assumption by considering the best matched box. For future WSOL researches, we encourage using \newmaxboxacc.}

{
\setlength{\tabcolsep}{5pt}
\renewcommand{\arraystretch}{0.9}
\begin{table}[t]
    \small
    \centering
    \begin{tabular}{*{2}{l}*{3}{r}}
         Statistics && ImageNet & \hspace{1em} CUB & OpenImages  \\
         \cline{1-1} \cline{3-5}
         \vspace{-.75em} & \\
         \#\ignorespaces Classes && $1000$ & $200$ & $100$ \\
         \vspace{-1em} & \\
         \#\ignorespaces images/class && \\
         \trainweaksup && $\sim\!1.2$K & $\sim\!30$ & $\sim\!300$ \\
         \trainfullsup && $10$ & $\sim\!5$ & $25$ \\
         \testfullsup && $10$ & $\sim\!29$ & $50$ \\
    \end{tabular}
    \caption{\small \textbf{Dataset statistics.} ``$\sim$'' indicates that the number of images per class varies across classes and the average value is shown.}
    \label{tab:dataset}
    \vspace{-1.5em}
\end{table}
}

\subsection{Data splits}
\label{subsec:evaluation_benchmarks}

For a fair comparison of the WSOL methods, we fix the amount of full supervision for hyperparameter search. As shown in Table~\ref{tab:dataset} we propose three disjoint splits for every dataset: \trainweaksup, \trainfullsup, and \testfullsup. The \trainweaksup contains images with weak supervision (the image-level labels). The \trainfullsup contains images with full supervision (either bounding box or binary mask). It is left as freedom for the user to utilize it for hyperparameter search, model selection, ablative studies, or even model fitting. The \testfullsup split contains images with full supervision; it must be used only for the final performance report. For example, checking the \testfullsup results multiple times with different model configurations violates the protocol as the learner implicitly uses more full supervision than allowed.

As WSOL benchmark datasets, ImageNet~\cite{ImageNet} and Caltech-UCSD Birds-200-2011 (CUB)~\cite{CUB} have been extensively used. 
For ImageNet, the $1.2$M ``train'' and $10$K ``validation'' images for $1\,000$ classes are treated as our \trainweaksup and \testfullsup, respectively. For \trainfullsup, we use the ImageNetV2~\cite{ImageNetV2}. We have annotated bounding boxes on those images. \firstround{For ImageNet we use the oracle box accuracy \boxacc.}

CUB has $5\,994$ ``train'' and $5\,794$ ``test'' images for 200 classes. We treat them as our \trainweaksup and \testfullsup, respectively. For \trainfullsup, we have collected $1\,000$ extra images ($\sim\!5$ images per class) from Flickr. \firstround{In addition, we have manually annotated bounding boxes. and automatically annotated foreground-background masks.}

We contribute a new WSOL benchmark based on the OpenImages instance segmentation subset~\cite{OpenImagesV5}. It provides a fresh WSOL benchmark to which the models have not yet overfitted. 
To balance the original OpenImages dataset, we have sub-sampled 100 classes and have randomly selected $29\,819$, $2\,500$, and $5\,000$ images from the original ``train'', ``validation'', and ``test'' splits as our \trainweaksup, \trainfullsup, and \testfullsup splits, respectively.
\firstround{For CUB and OpenImages we use the mask metric \mpxap.}
A summary of dataset statistics is in Table~\ref{tab:dataset}.

\addition{We summarize the following dataset contributions in this paper (contributions \textbf{bolded}):
\begin{itemize}
	\item CUB: \textbf{New data} (5 images $\times$ 200 classes) with \textbf{bounding box \firstround{and mask} annotations}.
	\item ImageNet: ImageNetV2~\cite{ImageNetV2} with new \textbf{bounding box annotations}.
	\item OpenImages: \textbf{Organized} the \trainweaksup, \trainfullsup, and \testfullsup splits for its use as a WSOL benchmark.
\end{itemize}

\subsubsection{ImageNet}
\label{appendixsub:imagenet}

The \textit{test set} of ImageNet-1k dataset~\cite{ImageNet} is not available. Therefore, many researchers report the accuracies on the \textit{validation set} for their final results~\cite{CutMix}. Since this practice may let models overfit to the evaluation split over time, ImageNetV2~\cite{ImageNetV2} has been proposed as the new test sets for ImageNet-1k trained models. 
We use the \texttt{Threshold0.7} split with $10\,000$ images (10 images per class) as our \trainfullsup. Since ImageNetV2 does not contain localization supervision, we have annotated $18\,532$ bounding boxes around each object.

\subsubsection{CUB}
\label{appendixsub:cub}

We have collected 5 images for each of the 200 CUB fine-grained bird classes from Flickr. The overall procedure is summarized as follows. Crawl images from Flickr; de-duplicate images against the original CUB dataset; manually prune irrelevant images (three people); prune with model classification scores; resize images; annotate bounding boxes. Sample images in Figure~\ref{fig:cub_v2_samples}. \firstround{For the automatically annotated masks, we have used the fully-supervised instance segmentation model, Cascade Mask R-CNN~\cite{MaskRCNN,CascadeMaskRCNN}, trained on MS-COCO dataset~\cite{COCO} with 3,362 bird samples to acquire segmentation masks for birds on the 1,000 CUB validation samples. We visualize random segmentation outputs in Figure~\ref{fig:cubv2_masks}; their qualities are close to human annotations (89.0 \pxap on the \testfullsup split against the ground-truth masks). }

\begin{figure}
    \centering
    \includegraphics[width=\linewidth]{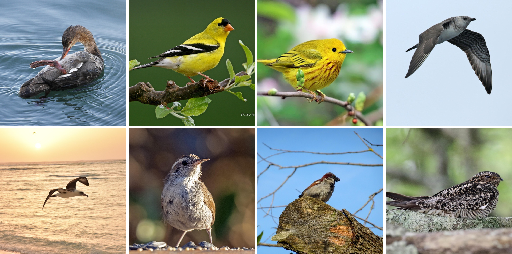}
    \caption{\small \addition{\textbf{CUB version 2.} Sample images.}}
    \label{fig:cub_v2_samples}
\end{figure}

\begin{figure}
    \centering
    \includegraphics[width=\linewidth]{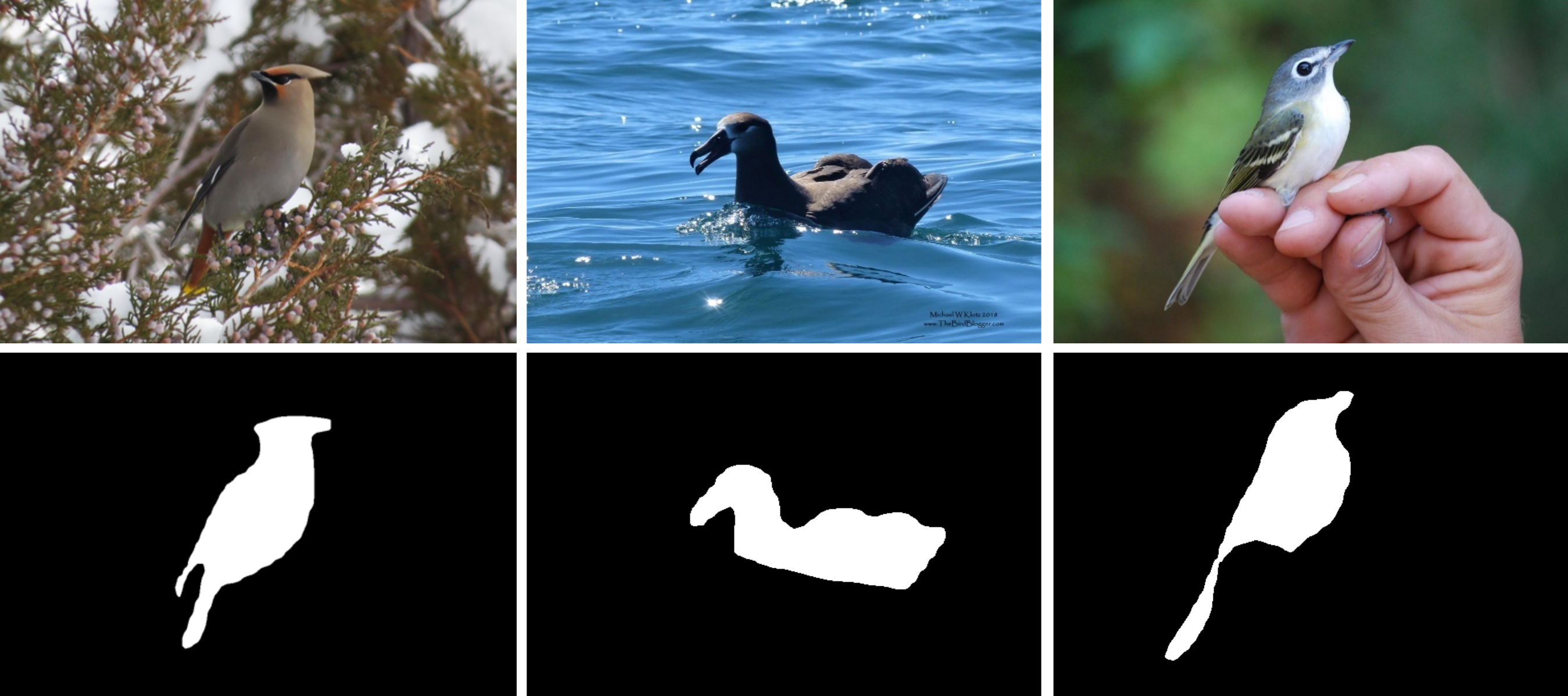}
    \caption{\small \firstround{\textbf{Masks for CUBV2.} We automatically annotate the CUBV2 dataset using Cascade Mask R-CNN~\cite{MaskRCNN,CascadeMaskRCNN}.}}
    \label{fig:cubv2_masks}
\end{figure}

\subsubsection{OpenImages}
\label{appendixsub:openimages}
There are three significant differences between OpenImagesV5~\cite{OpenImagesV5} and CUB or ImageNet that make the OpenImages not suitable as a WSOL benchmark in its original form. (1) Images are multi-labeled; it is not sensible to train classifiers with the standard softmax cross-entropy loss assuming single label per image. (2) OpenImages has less balanced label distributions. (3) There are nice instance segmentation masks, but they have many missing instances.

We have therefore processed a subset of OpenImages into a WSOL-friendly dataset where the above three issues are resolved. The procedure is as follows. Prune multi-labeled samples; exclude classes with not enough samples; randomly sample images for each class; prepare binary masks; introduce ignore regions. 
}

\subsection{Hyperparameter search}
\label{subsec:hyperparameter_search}

To make sure that the same amount of localization supervision is provided for each WSOL method, we refrain from employing any source of human prior outside the \trainfullsup split. 
If the optimal hyperparameter for an arbitrary dataset and architecture is not available by default, we subject it to the hyperparameter search algorithm.
For each hyperparameter, its \textit{feasible range}, as opposed to \textit{sensible range}, is used as the search space, to minimize the impact of human bias.

\definecolor{Gray}{gray}{0.85}
\newcolumntype{g}{>{\columncolor{Gray}}c}

{
	\setlength{\tabcolsep}{3pt}
	\begin{table*}[ht!]
		\centering
		\small
		\addition{\begin{tabular}{lc*{3}{c}gc*{3}{c}gc*{3}{c}gcg}
			& & \multicolumn{4}{c}{ImageNet} & & \multicolumn{4}{c}{CUB}  & & \multicolumn{4}{c}{OpenImages} && \multicolumn{1}{c}{Total}\\
			Methods  &  & VGG & Inception & ResNet & Mean &  & VGG & Inception & ResNet & Mean &  & VGG & Inception & ResNet & Mean &  & Mean \\
			\cline{1-1}\cline{3-6}\cline{8-11}\cline{13-16}\cline{18-18} & \vspace{-1em} \\
			CAM~\cite{CAM} &  & 0.936 & 0.949 & 0.922 & 0.936 &  & \firstround{0.951} & \firstround{0.945} & \firstround{0.913} & \firstround{0.936} &  & 0.966 & 0.922 & 0.926 & 0.938 &  & 0.927\\
			HaS~\cite{HaS} &  & 0.942 & 0.867 & 0.949 & 0.919 &  & \firstround{0.899} & \firstround{0.894} & \firstround{0.963} & \firstround{0.919} &  & 0.952 & 0.936 & 0.963 & 0.950 &  & 0.927\\
			ACoL~\cite{ACoL} &  & 0.968 & 1.000 & 0.950 & 0.973 &  & \firstround{0.988} & \firstround{0.924} & \firstround{0.984} & \firstround{0.965} &  & 0.930 & 0.959 & 0.952 & 0.947 &  & 0.938\\
			SPG~\cite{SPG} &  & 0.941 & 0.977 & 0.968 & 0.962 &  & \firstround{0.961} & \firstround{0.933} & \firstround{0.983} & \firstround{0.959} &  & 0.949 & 0.945 & 0.960 & 0.951 &  & 0.930\\
			ADL~\cite{ADL} &  & 0.945 & 0.954 & 0.995 & 0.965 &  & \firstround{0.941} & \firstround{0.903} & \firstround{0.945} & \firstround{0.930} &  & 0.957 & 0.936 & 0.913 & 0.935 &  & 0.917\\
			CutMix~\cite{CutMix} &  & 0.963 & 0.945 & 0.936 & 0.948 &  & \firstround{0.963} & \firstround{0.908} & \firstround{0.954} & \firstround{0.942} &  & 0.957 & 0.890 & 0.968 & 0.938 &  & 0.929\\
			\cline{1-1}\cline{3-6}\cline{8-11}\cline{13-16}\cline{18-18} & \vspace{-1em} \\
		\end{tabular}}
		\caption{\small \addition{\textbf{In-distribution ranking preservation.} Kendall's tau for the hyperparameter rankings between \trainfullsup and \testfullsup.}}
		\label{tab:supp-val-test-transfer}
	\end{table*}
}

We employ the random search hyperparameter optimization~\cite{RandomSearch}; it is simple, effective, and parallelizable. For each WSOL method, we sample 30 hyperparameters to train models on \trainweaksup and validate on \trainfullsup. The best hyperparameter combination is then selected.

\addition{To validate if the found hyperparameter rankings transfer well between the splits, we show the preservation of ranking statistics in Table~\ref{tab:supp-val-test-transfer}. We observe that the rankings are relatively well-preserved (with Kendall's tau values $>0.7$).}

\begin{figure}
    \centering
    \includegraphics[width=\linewidth]{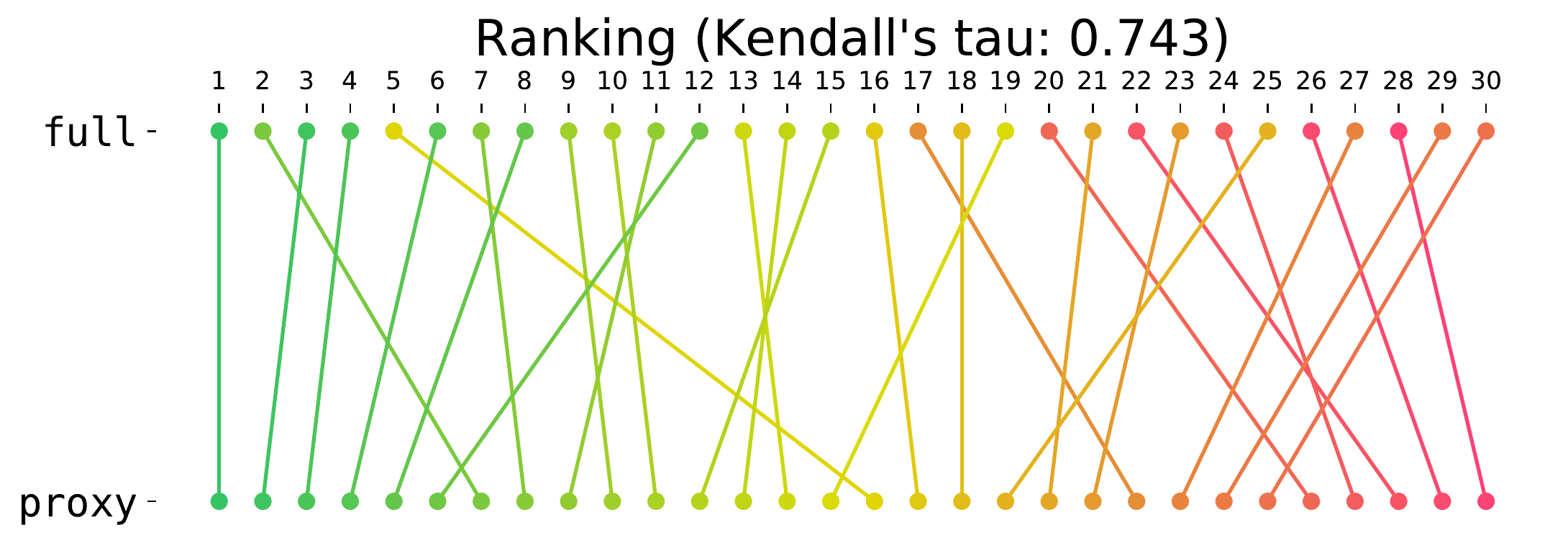}
    \caption{\small \addition{\textbf{Proxy ImageNet ranking}. Ranking of hyperparameters is largely preserved between the models trained on the full \trainweaksup and its 10\% proxy. Kendall's tau is $0.743$.}}
    \label{fig:proxy-imagenet}
\end{figure}

\revision{Since running 30 training sessions is costly for ImageNet ($1.2$M training images), we use 10\% of images in each class for fitting models during the search. We examine how much this reduction affects the rankings of hyperparameters; see Figure~\ref{fig:proxy-imagenet}. We observe again that the two rankings are largely preserved (Kendall's tau $0.743$).}

\section{Experiments}
\label{sec:results}

Based on the evaluation protocol in \S\ref{sec:evaluation}, we evaluate six previous weakly-supervised object localization (WSOL) methods (introduced in \S\ref{subsec:prior_wsol_methods}). We compare the performances (\S\ref{subsec:main_comparison_wsol}) and analyze the results (\S\ref{subsec:score_calibration_thresholding} and \S\ref{subsec:hyperparameter_analysis}). In addition to the above conference experiments, we provide experimental results for saliency methods, which may be considered methods for the WSOL task yet have seldom been evaluated as such (\S\ref{subsec:saliency_methods}). The analysis with few-shot learning (FSL) baselines has been updated since the conference version (\S\ref{subsec:few_shot_learning_results}), now with proper validation procedures.

\subsection{Evaluated methods}
\label{subsec:prior_wsol_methods}

We evaluate six widely used WSOL methods published in peer-reviewed venues. We describe each method in chronological order and discuss the set of hyperparameters. The full list of hyperparameters is in Table~\ref{tab:hyperparameter_list}.

\begin{table}[t]
	\footnotesize
	\centering
	\begin{tabular}{cccc}
		Methods && Hyperparameter & Distribution  \\
		\cline{1-1} \cline{3-4}
		\vspace{-1em} & \\
		\cline{1-1} \cline{3-4}
		\vspace{-1em} & \\
		Common && Learning rate & $\text{LogUniform}[10^{-5},10^0]$ \\
		&& Score-map resolution & $\text{Categorical}\{14,28\}$ \\
		\cline{1-1} \cline{3-4}
		\vspace{-1em} & \\
		HaS~\cite{HaS} && Drop rate & $\text{Uniform}[0,1]$ \\
		&& Drop area & $\text{Uniform}[0,1]$  \\
		\cline{1-1} \cline{3-4}
		\vspace{-1em} & \\  
		ACoL~\cite{ACoL} && Erasing threshold & $\text{Uniform}[0,1]$ \\
		\cline{1-1} \cline{3-4}
		\vspace{-1em} & \\
		SPG~\cite{SPG} && Threshold $\delta_{l}^{B1}$ & $\text{Uniform}[0,1]$ \\
		&& Threshold $\delta_{h}^{B1}$ & $\text{Uniform}[\delta_{l}^{B1},1]$ \\
		&& Threshold $\delta_{l}^{B2}$ & $\text{Uniform}[0,1]$ \\
		&& Threshold $\delta_{h}^{B2}$ & $\text{Uniform}[\delta_{l}^{B2},1]$ \\
		&& Threshold $\delta_{l}^{C}$ & $\text{Uniform}[0,1]$ \\
		&& Threshold $\delta_{h}^{C}$ & $\text{Uniform}[\delta_{l}^{C},1]$ \\
		\cline{1-1} \cline{3-4}
		\vspace{-1em} & \\
		ADL~\cite{ADL} && Drop rate & $\text{Uniform}[0,1]$ \\
		&& Erasing threshold & $\text{Uniform}[0,1]$ \\
		\cline{1-1} \cline{3-4}
		\vspace{-1em} & \\
		CutMix~\cite{CutMix} && Size prior & $\frac{1}{\text{Uniform}(0,2]}-\frac{1}{2}$ \\
		&& Mix rate & $\text{Uniform}[0,1]$ \\
		\cline{1-1} \cline{3-4}
		\vspace{-1em} & \\
		\multirow{2}{*}{GC-Net} &  & $\mathcal{L}_\text{area}$       & Categorical\{Yes, No\} \\
                        &  & $\mathcal{L}_\text{background}$ & Categorical\{Yes, No\} \\
		\cline{1-1} \cline{3-4}
	\end{tabular}
	\vspace{1em}
	\caption{\small \textbf{Hyperparameter search spaces.} }
	\label{tab:hyperparameter_list}
\end{table}

\noindent
\textbf{Class activation mapping (CAM)}~\cite{CAM} trains a classifier of fully-convolutional backbone with the global average pooling structure. At test time, CAM uses the logit outputs before GAP as the score map $s_{ij}$. CAM has the learning rate and the score-map resolution as hyperparameters and all five methods below use CAM in the background. \addition{learning rate is sampled log-uniformly from $[10^{-5},10^{0}]$, where end points correspond roughly to ``no training'' and ``training always diverges'' cases. Score-map resolution is sampled from $\text{Categorical}\{14,28\}$, two widely used resolutions in prior WSOL methods. All five methods below use CAM technique in the background, and have learning rate and score-map resolution as design choices.}

\noindent
\revision{\textbf{Hide-and-seek (HaS)}~\cite{HaS} is a data augmentation technique that divides an input image into grid-like patches, and then randomly select patches to be dropped.} \addition{The hyperparameters of HaS are drop rate and drop area. Specifically, the size of each patch is decided by drop area, and the probability of each patch to be selected for erasing is decided by drop rate. Drop area is sampled from a uniform distribution $U[0,1]$, where $0$ corresponds to ``no grid'' and $1$ indicates ``full image as one patch''.}

\noindent
\textbf{Adversarial complementary learning (ACoL)}~\cite{ACoL} proposes an architectural solution: a two-head architecture where one adversarially erases the high-scoring activations in the other. \addition{From one head, ACoL finds the high-score region using CAM and erases it from an internal feature map. The other head learns remaining regions using the erased feature map. We sample erasing threshold from a uniform distribution $U[0,1]$, where $0$ means ``erasing whole feature map'' and $1$ means ``do not erase''.}

\noindent
\textbf{Self-produced guidance (SPG)}~\cite{SPG} is another architectural solution where internal pseudo-pixel-wise supervision is synthesized on the fly. \addition{SPG utilizes spatial information about fore- and background using three additional branches (\texttt{SPG-B1,B2,C}). To divide foreground and background from score-map, they introduce two hyperparameters, $\delta_{l}$ and $\delta_{h}$, per each branch. When the score is lower than $\delta_{l}$, the pixel is considered as background, and the pixel is considered as foreground when the score is higher than  $\delta_{h}$. The remaining region (higher than $\delta_{l}$, lower than  $\delta_{h}$) is ignored. We first sample $\delta_{l}$ from $U[0,1]$, and then $\delta_{h}$ is sampled from $U[\delta_{l},1]$.}

\noindent
\textbf{Attention-based dropout layer (ADL)}~\cite{ADL} has proposed a module that, like ACoL, adversarially produces drop masks at high-scoring regions, while not requiring an additional head.
\addition{ADL produces a drop mask by finding the high-score region to be dropped using another scoring rule~\cite{zagoruyko2017paying}. Also, ADL produces an importance map by normalizing the score map and uses it to increase classification power of the backbone. At each iteration, only one component is applied between the drop mask and importance map. The hyperparameters of ADL are drop rate that indicates how frequently the drop mask is selected and erasing threshold that means how large regions are dropped. We sample both hyperparameters from uniform distributions $U[0,1]$.}

\noindent
\revision{\textbf{CutMix}~\cite{CutMix} is a data augmentation technique, where patches in training images are cut and pasted to other images and target labels are mixed likewise.} \addition{Its hyperparameters consist of the size prior $\beta$ (used for sampling sizes according to $\sim\!\!\text{Beta}(\beta,\beta)$) and the mix rate $r$ (Bernoulli decision for ``CutMix or not''). The size prior is sampled from the positive range $\frac{1}{\text{Unif}(0,2]}-\frac{1}{2}$; then, $\text{Var}(\text{Beta}(\beta,\beta))$ follows the uniform distribution between 0 and 0.25 (maximal variance; two Dirac deltas at 0 and 1).}

\noindent
\firstround{\textbf{Geometry Constrained Network (GC-Net)}~\cite{GCNet} is a WSOL method of different nature. Unlike other WSOL methods that predict object locations as pixel-wise score maps, GC-Net predicts parametrized shapes (\eg rectangles and ellipses) that tightly contain the foreground object.
The tightness of the shapes are regularized by the area loss $\mathcal{L}_\text{area}$ that penalizes shapes of greater areas. To make sure the shapes fully cover the object area, background loss $\mathcal{L}_\text{background}$ has been proposed to penalize confident predictions on the image where the shape is cropped out; it tends to enlarge the shape.
}

\myparagraph{Few-shot learning (FSL) baseline.}
The full supervision in \trainfullsup used for validating WSOL hyperparameters can be used for training a model itself. Since only a few fully labeled samples per class are available, we refer to this setting as the few-shot learning (FSL) baseline.

As a simple baseline, we consider a foreground saliency mask predictor~\cite{FirstSaliency}. We alter the last layer of a fully convolutional network (FCN) into a $1\times 1$ convolutional layer with $H\times W$ score map output. Each pixel is trained with the binary cross-entropy loss against the target mask, as done in~\cite{Deeplab,FCN,oh2017exploiting}. For OpenImages, the pixel-wise masks are used as targets; for ImageNet and CUB, we build the mask targets by labeling pixels inside the ground truth boxes as foreground~\cite{BoxWSSS}. At inference phase, the $H\times W$ score maps are evaluated with the box or mask metrics.

\myparagraph{Center-gaussian baseline.}
The Center-gaussian baseline generates isotropic Gaussian score maps centered at the images. We set the standard deviation to 1, but note that it does not affect the \maxboxacc and \pxap measures. This provides a no-learning baseline for every localization method.

\revision{\myparagraph{Which checkpoint is suitable for evaluation?} 
We observe in our preliminary experiments that, unlike for classification performances, localization performances go through significant amount of fluctuations in the earlier epochs, resulting in unstable maximal performances. We thus compare the last checkpoints from each method, after the training has sufficiently converged. We recommend following this practice for future WSOL researchers.}

\firstround{
\begin{table*}
\centering
{
\footnotesize
\setlength{\tabcolsep}{0.30em}
\firstround{\begin{tabular}{l*{22}{c}}
\vspace{-1em} & \\
&&\hspace{0.5em} & \multicolumn{3}{c}{ImageNet} & \hspace{0.3em} & \multicolumn{3}{c}{CUB} & \hspace{0.5em} & \multicolumn{3}{c}{ImageNet} & \hspace{0.3em} & \multicolumn{3}{c}{CUB} & \hspace{0.3em} & \multicolumn{3}{c}{OpenImages}  \\
&&\hspace{0.5em} & \multicolumn{3}{c}{\texttt{Top-1 loc}} & \hspace{0.3em} & \multicolumn{3}{c}{\texttt{Top-1 loc}} & \hspace{0.5em} & \multicolumn{3}{c}{\maxboxacc} & \hspace{0.3em} & \multicolumn{3}{c}{\maxboxacc} & \hspace{0.3em} & \multicolumn{3}{c}{\mpxap}  \\
\cline{4-6} \cline{8-10} \cline{12-14} \cline{16-18} \cline{20-22}
\vspace{-1em} & \\
&Methods && {V} &{I} &{R} &&{V} &{I} &{R} &&{V} &{I} &{R} &&{V} &{I} &{R} &&{V} &{I} &{R} \\
\cline{2-2} \cline{4-6} \cline{8-10} \cline{12-14} \cline{16-18} \cline{20-22}
\vspace{-1em} & \\
\cline{2-2} \cline{4-6} \cline{8-10} \cline{12-14} \cline{16-18} \cline{20-22}
\vspace{-1em} & \\
\multirow{7}{*}{\rotatebox{90}{Reported\hspace{0.0em}}} & CAM~\cite{CAM} &  & 42.8 & - & 46.3 &  & 37.1 & 43.7 & 49.4 &  & - & 62.7 & - &  & - & - & - &  & - & - & -\\
 & HaS~\cite{HaS} &  & - & - & - &  & - & - & - &  & - & - & - &  & - & - & - &  & - & - & -\\
 & ACoL~\cite{ACoL} &  & 45.8 & - & - &  & 45.9 & - & - &  & - & - & - &  & - & - & - &  & - & - & -\\
 & SPG~\cite{SPG} &  & - & 48.6 & - &  & - & 46.6 & - &  & - & 64.7 & - &  & - & - & - &  & - & - & -\\
 & ADL~\cite{ADL} &  & 44.9 & 48.7 & - &  & 52.4 & 53.0 & - &  & - & - & - &  & 75.4 & - & - &  & - & - & -\\
 & CutMix~\cite{CutMix} &  & 43.5 & - & 47.3 &  & - & 52.5 & 54.8 &  & - & - & - &  & - & - & - &  & - & - & -\\
 & GC-Net~\cite{GCNet} &  & - & 49.1 & - &  & 58.9 & - & - &  & - & - & - &  & 74.9 & - & - &  & - & - & -\\
\cline{2-2} \cline{4-6} \cline{8-10} \cline{12-14} \cline{16-18} \cline{20-22}
\vspace{-1em} & \\
\multirow{7}{*}{\rotatebox{90}{Reproduced\hspace{0.0em}}} & CAM~\cite{CAM} &  & 45.5 & 48.8 & 51.8 &  & 45.8 & 40.4 & 56.1 &  & 61.1 & 65.3 & 64.2 &  & 71.1 & 62.1 & 73.2 &  & 58.3 & 63.2 & 58.5\\
 & HaS~\cite{HaS} &  & 46.3 & 49.7 & 49.9 &  & 55.6 & 41.1 & 60.7 &  & 61.9 & 65.5 & 63.1 &  & 76.2 & 57.7 & 78.1 &  & 58.1 & 58.1 & 55.9\\
 & ACoL~\cite{ACoL} &  & 45.5 & 49.9 & 47.4 &  & 44.8 & 46.8 & 57.8 &  & 60.3 & 64.6 & 61.6 &  & 72.3 & 59.5 & 72.7 &  & 54.3 & 57.2 & 57.3\\
 & SPG~\cite{SPG} &  & 44.6 & 48.6 & 48.5 &  & 42.9 & 44.9 & 51.5 &  & 61.6 & 65.5 & 63.4 &  & 63.7 & 62.7 & 71.4 &  & 58.3 & 62.3 & 56.7\\
 & ADL~\cite{ADL} &  & 44.4 & 45.0 & 51.5 &  & 39.2 & 35.2 & 41.1 &  & 60.8 & 61.6 & 64.1 &  & 75.6 & 63.3 & 73.5 &  & 58.1 & 62.6 & 57.7\\
 & CutMix~\cite{CutMix} &  & 46.1 & 49.2 & 51.5 &  & 47.0 & 48.3 & 54.5 &  & 63.9 & 62.2 & 65.4 &  & 71.9 & 65.5 & 67.8 &  & 58.7 & 63.2 & 58.5\\
 & GC-Net~\cite{GCNet} &  & - & - & - &  & 59.3 & - & - &  & - & - & - &  & 74.1 & - & - &  & - & - & -\\
\cline{2-2} \cline{4-6} \cline{8-10} \cline{12-14} \cline{16-18} \cline{20-22}
\end{tabular}
}
{
\small
\setlength{\tabcolsep}{0.4em}
\firstround{\begin{tabular}{cll}
     & \\
     &&Architecture  \\
     \cline{1-1}\cline{3-3}
     V&&VGG-GAP~\cite{VGG}\\
     I&&InceptionV3~\cite{InceptionV3}\\
     R&&ResNet50~\cite{ResNet} \\
     \cline{1-1}\cline{3-3}
\vspace{-13.2em}
\end{tabular}}
}
}
\caption{\small \firstround{\textbf{Previously reported WSOL results.} The first six rows are reported results in prior WSOL papers. When there are different performance reports for the same method in different papers, we choose the greater performance.}}
\label{tab:previous_current_results}
\end{table*}
}

\definecolor{darkergreen}{RGB}{21, 152, 56}
\definecolor{red2}{RGB}{252, 54, 65}
\definecolor{Gray}{gray}{0.85}
\newcolumntype{g}{>{\columncolor{Gray}}c}
\newcommand\tableminus[1]{\textcolor{red2}{#1}}
\newcommand\tableplus[1]{\textcolor{darkergreen}{#1}}

{
	\setlength{\tabcolsep}{3pt}
	\renewcommand{\arraystretch}{1.1}
	\begin{table*}[ht!]
		\resizebox{\textwidth}{!}{%
			\centering
			\small
			\begin{tabular}{lc*{3}{c}gc*{3}{c}gc*{3}{c}gcg}
				& & \multicolumn{4}{c}{ImageNet (\newmaxboxacc)} & & \multicolumn{4}{c}{CUB (\mpxap)}  & & \multicolumn{4}{c}{OpenImages (\mpxap)} && \multicolumn{1}{c}{Total}\\
				Methods  &  & VGG & Inception & ResNet & Mean &  & VGG & Inception & ResNet & Mean &  & VGG & Inception & ResNet & Mean &  & Mean \\
				\cline{1-1}\cline{3-6}\cline{8-11}\cline{13-16}\cline{18-18} & \vspace{-1em} \\
				CAM~\cite{CAM} &  & 60.0 & 63.4 & 63.7 & 62.4 &  & 75.4 & 70.4 & 66.6 & 70.8 &  & 59.2 & 63.6 & 58.7 & 60.5 &  & 64.5\\
				HaS~\cite{HaS} &  & \tableplus{+0.6} & \tableplus{+0.3} & \tableminus{-0.3} & \tableplus{+0.2} &  & \tableminus{-4.2} & \tableminus{-4.9} & \tableplus{+3.5} & \tableminus{-1.9} &  & \tableminus{-0.2} & \tableminus{-4.1} & \tableminus{-2.9} & \tableminus{-2.4} &  & \tableminus{-1.4}\\
				ACoL~\cite{ACoL} &  & \tableminus{-2.6} & \tableplus{+0.3} & \tableminus{-1.4} & \tableminus{-1.2} &  & \tableminus{-10.0} & \tableminus{-6.7} & \tableplus{+0.1} & \tableminus{-5.5} &  & \tableminus{-5.2} & \tableminus{-7.1} & \tableminus{-0.4} & \tableminus{-4.2} &  & \tableminus{-3.7}\\
				SPG~\cite{SPG} &  & \tableminus{-0.1} & \tableminus{-0.1} & \tableminus{-0.4} & \tableminus{-0.2} &  & \tableminus{-7.5} & \tableplus{+0.3} & \tableplus{+3.1} & \tableminus{-1.4} &  & \tableminus{-0.6} & \tableminus{-0.7} & \tableminus{-1.8} & \tableminus{-1.0} &  & \tableminus{-0.9}\\
				ADL~\cite{ADL} &  & \tableminus{-0.2} & \tableminus{-2.0} & \tableplus{+0.0} & \tableminus{-0.7} &  & \tableplus{+1.9} & \tableminus{-1.2} & \tableminus{-4.4} & \tableminus{-1.2} &  & \tableminus{-0.4} & \tableminus{-6.6} & \tableminus{-3.0} & \tableminus{-3.3} &  & \tableminus{-1.8}\\
				CutMix~\cite{CutMix} &  & \tableminus{-0.6} & \tableplus{+0.5} & \tableminus{-0.4} & \tableminus{-0.2} &  & \tableminus{-2.1} & \tableminus{-1.6} & \tableplus{+0.8} & \tableminus{-1.0} &  & \tableminus{-0.4} & \tableminus{-0.4} & \tableminus{-0.3} & \tableminus{-0.4} &  & \tableminus{-0.5}\\
				\firstround{GC-Net~\cite{GCNet}} &  & \tableminus{-5.6} & \tableminus{-15.1} & \tableminus{-8.9} & \tableminus{-9.9} &  & - & - & - & - &  & - & - & - & - &  & - \\
				\cline{1-1}\cline{3-6}\cline{8-11}\cline{13-16}\cline{18-18} & \vspace{-1em} \\
				Best WSOL &  & 60.6 & 63.9 & 63.7 & 62.6 &  & 77.3 & 70.7 & 70.1 & 70.8 &  & 59.2 & 63.6 & 58.7 & 60.5 &  & 64.5\\
				FSL baseline &  & 61.6 & 68.8 & 66.3 & 65.6 &  & 76.6 & 89.2 & 89.1 & 85.0 &  & 60.0 & 71.0 & 68.3 & 66.4 &  & 72.3\\
				Center baseline &  & 48.9 & 48.9 & 48.9 & 48.9 &  & 55.9 & 55.9 & 55.9 & 55.9 &  & 46.7 & 46.7 & 46.7 & 46.7 &  & 50.5\\
				\cline{1-1}\cline{3-6}\cline{8-11}\cline{13-16}\cline{18-18} & \vspace{-1em} \\
			\end{tabular}
		}
		\caption{\small \textbf{Re-evaluating WSOL.} How much have WSOL methods improved upon the vanilla CAM model? \testfullsup split results are shown, relative to the vanilla CAM performance (\tableplus{increase} or \tableminus{decrease}). Hyperparameters have been optimized over the identical \trainfullsup split for all WSOL methods and the FSL baseline: (10,5,25) full supervision/class for (ImageNet,CUB,OpenImages).} 
		\label{tab:main_v2}
	\end{table*}
}

\definecolor{darkergreen}{RGB}{21, 152, 56}
\definecolor{red2}{RGB}{252, 54, 65}
\definecolor{Gray}{gray}{0.85}
\newcolumntype{g}{>{\columncolor{Gray}}c}

{
\small
	\setlength{\tabcolsep}{3pt}
	\renewcommand{\arraystretch}{1.1}
	\begin{table*}[ht!]
		\resizebox{\textwidth}{!}{
			\centering
			\small
			\addition{\begin{tabular}{lc*{3}{c}gc*{3}{c}gc*{3}{c}gcg}
				& & \multicolumn{4}{c}{ImageNet} & & \multicolumn{4}{c}{CUB}  & & \multicolumn{4}{c}{OpenImages} && \multicolumn{1}{c}{Total}\\
				Methods  &  & VGG & Inception & ResNet & Mean &  & VGG & Inception & ResNet & Mean &  & VGG & Inception & ResNet & Mean &  & Mean \\
				\cline{1-1}\cline{3-6}\cline{8-11}\cline{13-16}\cline{18-18} & \vspace{-1em} \\
				CAM~\cite{CAM}  &    &  66.5  &  70.6  &  75.0  &  70.7  &    &  50.1  &  70.7  &  71.5  &  64.1  &    &  70.2  &  56.9  &  74.5  &  67.2  &    &  67.3\\
				HaS~\cite{HaS}  &    &  68.3  &  69.1  &  75.4  &  70.9  &    &  75.9  &  64.5  &  69.7  &  70.0  &    &  68.3  &  66.2  &  73.8  &  69.4  &    &  70.1\\
				ACoL~\cite{ACoL}  &    &  64.5  &  71.8  &  73.1  &  69.8  &    &  71.8  &  71.5  &  71.1  &  71.4  &    &  70.2  &  61.9  &  70.8  &  67.6  &    &  69.6\\
				SPG~\cite{SPG}  &    &  67.8  &  71.1  &  73.3  &  70.7  &    &  72.1  &  46.2  &  50.4  &  56.3  &    &  66.8  &  69.0  &  70.8  &  68.9  &    &  65.3\\
				ADL~\cite{ADL}  &    &  67.6  &  61.2  &  72.0  &  66.9  &    &  55.0  &  41.0  &  66.6  &  54.2  &    &  68.5  &  63.0  &  62.9  &  64.8  &    &  62.0\\
				CutMix~\cite{CutMix}  &    &  66.4  &  69.2  &  75.7  &  70.4  &    &  48.4  &  71.0  &  73.0  &  64.1  &    &  69.6  &  54.4  &  74.1  &  66.0  &    &  66.9\\
				\firstround{GC-Net~\cite{GCNet}}  &    &  \firstround{70.0}  &  \firstround{67.3}  &  \firstround{74.5}  &  \firstround{70.7}  &    &  -  &  -  &  -  &  -  &    &  -  &  -  &  -  &  -  &    &  -\\
				\cline{1-1}\cline{3-6}\cline{8-11}\cline{13-16}\cline{18-18} & \vspace{-1em} \\
			\end{tabular}}
		}
		\caption{\small \addition{\textbf{Classification performance of WSOL methods.} Classification accuracies of the models in Table~\ref{tab:main_v2}. Hyperparameters for each model are optimal for the localization performances on \trainfullsup split; they may be sub-optimal for classification accuracies.}}
		\label{tab:main_cls}
	\end{table*}
}

\subsection{Comparison of WSOL methods}
\label{subsec:main_comparison_wsol}
We evaluate the six WSOL methods over three backbone architectures, \ie VGG-GAP~\cite{VGG,CAM}, InceptionV3~\cite{InceptionV3}, and ResNet50~\cite{ResNet}, and three datasets, \ie CUB, ImageNet and OpenImages. For each (method, backbone, dataset) tuple, we have randomly searched the optimal hyperparameters over the \trainfullsup with 30 trials, totalling about $9\,000$ GPU hours. Since the sessions are parallelizable, it has taken only about 200 hours over 50 P40 GPUs to obtain the results. The results are shown in Table~\ref{tab:main_v2}. We use the same batch sizes and training epochs to enforce the same computational budget. \revision{The last checkpoints are used for evaluation.}

\addition{
\myparagraph{Reported performances for prior WSOL methods.}
Before studying the unified evaluation, we examine the reported progresses in the WSOL task in previous papers. The numbers are summarized in Table~\ref{tab:previous_current_results}. The score reports indicate a strong trend for improvement in localization scores (both in top-1 and GT-known localization metrics). For example, the (GT-known, ImageNet, GoogleNet) case shows an improvement from 58.7 (CAM; 2016) to 60.6 (HaS; 2017) and 63.0 (ACoL; 2018). At the same time, we observe that the metrics, datasets, and architectures have not been unified in those papers; every paper since CAM has considered a hardly overlapping set of architecture-dataset pair against the prior arts. Our paper prepares a ground for comparing WSOL method on the same set of architecture-dataset pairs with the rectified evaluation protocols and metrics.
}

\firstround{
\myparagraph{Reproducibility of our implementation.} Table~\ref{tab:previous_current_results} summarizes the reported performances and our re-implemented results for each method on each dataset-architecture pair. Note that our re-implementations have reproduced the results generally well, with often better performances than previously reported (e.g. 62.7\% to 65.3\% for CAM on ImageNet with InceptionV3). One exception is ADL with the reported result of 52.4\% top-1 localization accuracy on CUB with VGG backbone; our re-implementation results in 39.2\%. This is attributable to the reduced training epochs for our unified, fair training setup.
}

\myparagraph{Comparison under unified evaluation framework.}
The results are shown in Table~\ref{tab:main_v2}. WSOL methods have actually not improved significantly since CAM~\cite{CAM}, when validated in the same data splits and same evaluation metrics. \revision{On ImageNet, methods after CAM are generally struggling: only HaS has seen a boost of +0.2pp on average.} \firstround{We observe that GC-Net surpasses the center baseline, but has significantly worse localization performances than CAM and other WSOL methods.} \revision{On CUB and the new WSOL benchmark, OpenImages, no method has improved over CAM.} In general, we observe a random mixture of increases and decreases in performance over the baseline CAM, depending on the architecture and dataset. Overall, CAM achieves the best averaged performance of \firstround{64.5\%}. An important result in the table to be discussed later is the comparison against the few-shot learning baseline (\S\ref{subsec:few_shot_learning_results}).

\myparagraph{Why are there discrepancies?}
There are many reasons for the differences in the conclusions between the previous reported results (Table~\ref{tab:previous_current_results}) and our re-evaluations (Table~\ref{tab:main_v2}). (1) Our evaluation metric is based on GT-known localization performances, while many prior papers have adopted the top-1 localization accuracies that confound the classification and localization performances. (2) We resolve another confounding factor: the boost from the actual score map improvement and that from the score normalization and thresholding. We make our evaluation independent of the latter. (3) Different types and amounts of full supervision employed under the hood; we assign the same number of fully-supervised validation samples per method. (4) The use of different training settings (\eg batch sizes and epochs). Since those settings are not published for many WSOL methods, we decide to match the training budget for fair comparisons: training epochs are ($10$, $50$, $10$) for (ImageNet, CUB, OpenImages) and the batch size is always $32$.

\addition{\myparagraph{Classification results of WSOL models.}
We do not report the widely-used ``top-1 localization accuracy'', as it confounds the classification \textit{and} localization performances. We suggest the ``GT-known'' type of metrics like \maxboxacc and \pxap that measures the localization performances given perfect classification. We separately report the classification performances in Table~\ref{tab:main_cls}, for a complete analysis. 

Unlike localization performances, HaS and ACoL improves the classification performances over CAM (+2.8pp and +2.3pp total mean accuracies, respectively).  \firstround{We argue that these methods should be separately acknowledged as valuable regularization methods for classifiers.} 
We observe in general that the classification performances do not correlate with the localization performances. The apparent improvements shown in previous papers in terms of the ``top-1 localization scores'' may partly be explained by the improvements in classification performances. The result signifies that the two performances must be separately measured.}

\subsection{Score calibration and thresholding}
\label{subsec:score_calibration_thresholding}

{
\setlength{\tabcolsep}{3pt}
\renewcommand{\arraystretch}{1.1}
\begin{table*}[ht!]
	\resizebox{\textwidth}{!}{%
		\centering
		\small
		{\begin{tabular}{lc*{3}{c}gc*{3}{c}gc*{3}{c}gcg}
			& & \multicolumn{4}{c}{ImageNet (\newmaxboxacc)} & & \multicolumn{4}{c}{CUB (\mpxap)}  & & \multicolumn{4}{c}{OpenImages (\mpxap)} && \multicolumn{1}{c}{Total}\\
			Methods  &  & VGG & Inception & ResNet & Mean &  & VGG & Inception & ResNet & Mean &  & VGG & Inception & ResNet & Mean &  & Mean \\
			\cline{1-1}\cline{3-6}\cline{8-11}\cline{13-16}\cline{18-18} & \vspace{-1em} \\
			CAM~\cite{CAM,GradCAM}	& &	60.9	&	64.8	&	{64.9}	&	63.5 	& &	76.2	&	72.9	&	67.7	&	72.3	& &	{60.0}	&	{64.4}	&	58.8	&	{61.2}	& &	65.7	\\
            HaS~\cite{HaS}	& &	\tableminus{-1.4}	&	\tableminus{-1.8}	&	\tableminus{-0.2}	&	\tableminus{-1.1}	& &	\tableplus{+0.9}	&	\tableminus{-2.8}	&	\tableplus{+3.6}	&	\tableplus{+0.5}	& &	\tableminus{-0.4}	&	\tableminus{-2.6}	&	\tableminus{-3.3}	&	\tableminus{-2.1}	& &	\tableminus{-0.9} 	\\
            ACoL~\cite{ACoL}	& &	\tableminus{-5.2}	&	\tableminus{-2.8}	&	\tableminus{-4.7}	&	\tableminus{-4.2}	& &	\tableminus{-10.8}	&	\tableminus{-9.2}	&	\tableminus{-1.0}	&\tableminus{-7.0}	& &	\tableminus{-6.0}	&	\tableminus{-7.9}	&	\tableminus{-1.0}	&	\tableminus{-4.9}	& &	\tableminus{-5.4} \\
            SPG~\cite{SPG}	& &	\tableminus{-0.5}	&	\tableminus{-0.2}	&	\tableminus{-0.8}	&	\tableminus{-0.5}	& &	\tableminus{-2.6}	&	\tableplus{+0.2}	&	\tableplus{+3.0}	&	\tableminus{+0.2}	& &	\tableminus{-0.7}	&	\tableplus{+0.0}	&	\tableminus{-2.2}	&	\tableminus{-0.9}	& &	\tableminus{-0.4} 	\\
            ADL~\cite{ADL}	& &	\tableplus{+2.3}	&	\tableminus{-1.9}	&	\tableminus{-0.3}	&	\tableplus{+0.1}	& &	\tableplus{+1.9}	&	\tableminus{-1.1}	&	\tableminus{-4.4}	&	\tableminus{-1.2}	& &	\tableminus{-0.6}	&	\tableminus{-6.9}	&	\tableminus{-3.6}	&	\tableminus{-3.7}	& &	\tableminus{-1.6} \\
            CutMix~\cite{CutMix}	& &	\tableminus{-0.6}	&	\tableplus{+0.6}	&	\tableminus{-0.3}	&	\tableminus{-0.1}	& &	\tableplus{+3.3}	&	\tableminus{-0.8}	&	\tableplus{+0.8}	&	\tableplus{+1.1}	& &	\tableminus{-0.5}	&	\tableminus{-0.2}	&	\tableplus{+0.1}	&	\tableminus{-0.2}	& &	\tableplus{+0.2} \\
			\cline{1-1}\cline{3-6}\cline{8-11}\cline{13-16}\cline{18-18} & \vspace{-1em} \\
			Best WSOL &  & 63.1 & 64.9 & 64.9 & 63.6 &  & 79.5 & 73.1 & 71.3 & 73.4 &  & 60.0 & 64.4 & 59.4 & 61.2 &  & 65.9\\
			FSL baseline &  & 61.6 & 68.8 & 66.3 & 65.6 &  & 76.6 & 89.2 & 89.1 & 85.0 &  & 60.0 & 71.0 & 68.3 & 66.4 &  & 72.3\\
			Center baseline &  & 48.9 & 48.9 & 48.9 & 48.9 &  & 55.9 & 55.9 & 55.9 & 55.9 &  & 46.7 & 46.7 & 46.7 & 46.7 &  & 50.5\\
            \cline{1-1}\cline{3-6}\cline{8-11}\cline{13-16}\cline{18-18}\\
		\end{tabular}
	}}
	\caption{\small \firstround{\textbf{WSOL results with max normalization.} While we focus on min-max normalization throughout the paper, we show the results with max normalization. As in Table~\ref{tab:main_v2}, we show the relative performances against CAM. } }
	\label{tab:gradcam}
\end{table*}
}

\begin{figure}
	\centering
	\includegraphics[width=0.93\linewidth]{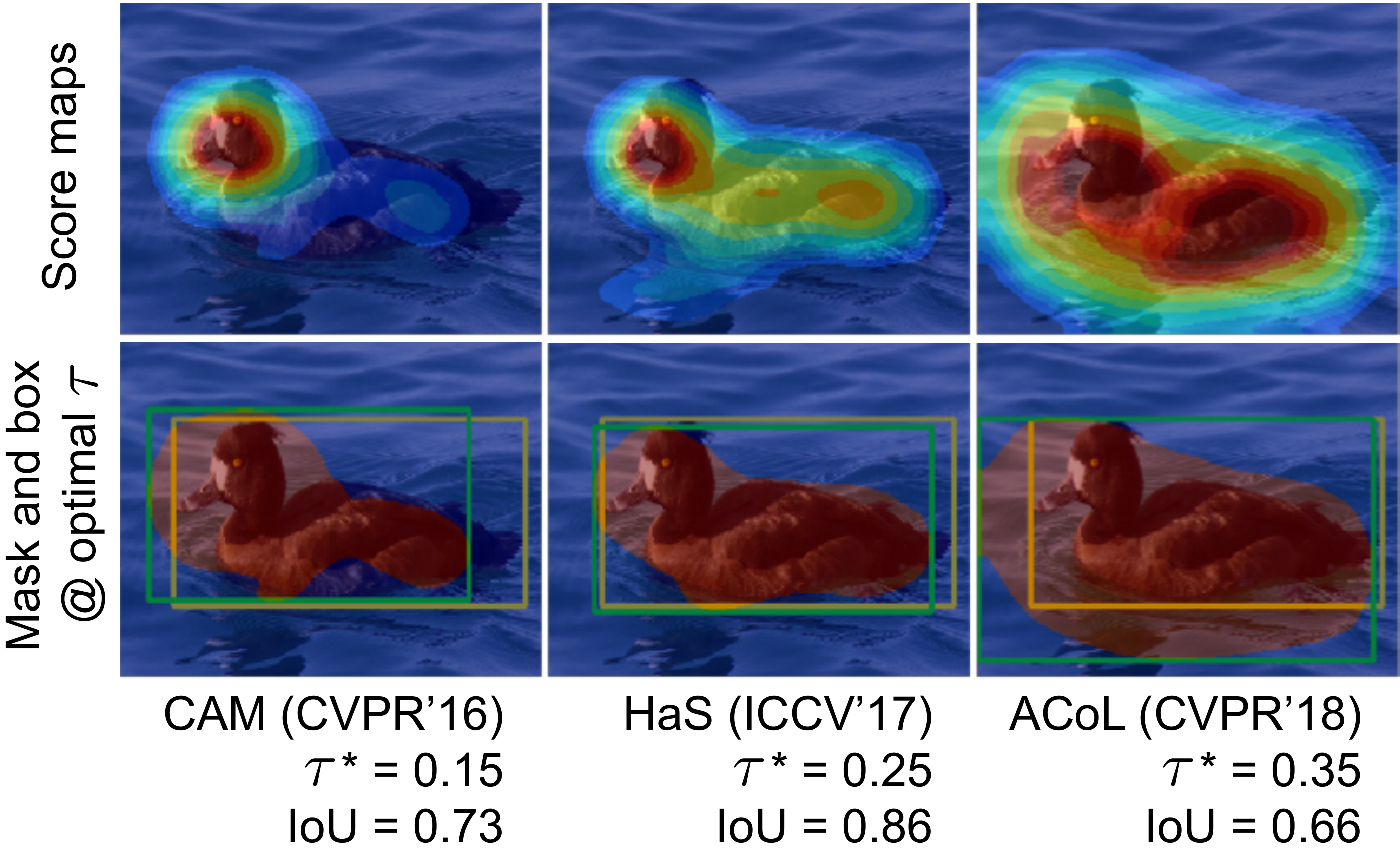}
	\caption{\small \textbf{Selecting $\tau$.} Measuring performance at a fixed threshold $\tau$ can lead to a false sense of improvement. Compared to CAM, HaS and ACoL expand the score maps, but they do not necessarily improve the box qualities (IoU) at the optimal $\tau^\star$. Predicted and ground-truth boxes are shown as green and yellow boxes.}
	\label{fig:qualitative_cam_threshold}
\end{figure}

\begin{figure*}
	\centering
	\begin{subfigure}[b]{\linewidth}
		\includegraphics[width=\linewidth]{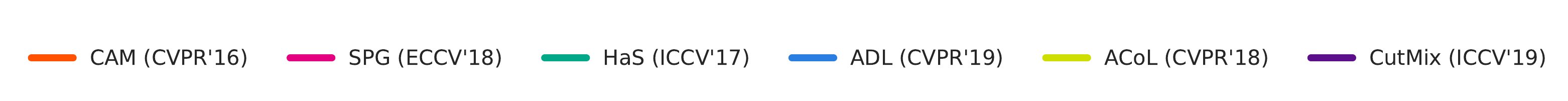}
	\end{subfigure}
	
	\begin{subfigure}[b]{.31\linewidth}
		\includegraphics[width=\linewidth]{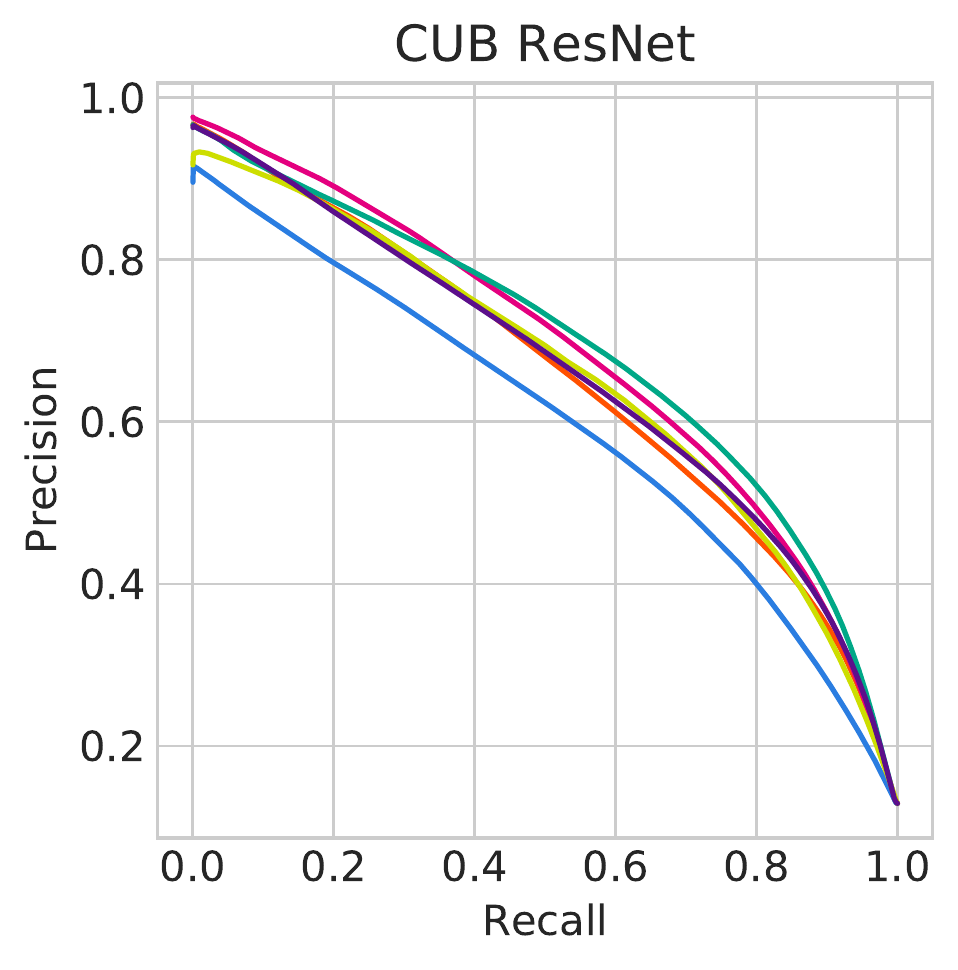}
	\end{subfigure}
	\begin{subfigure}[b]{.31\linewidth}
		\includegraphics[width=\linewidth]{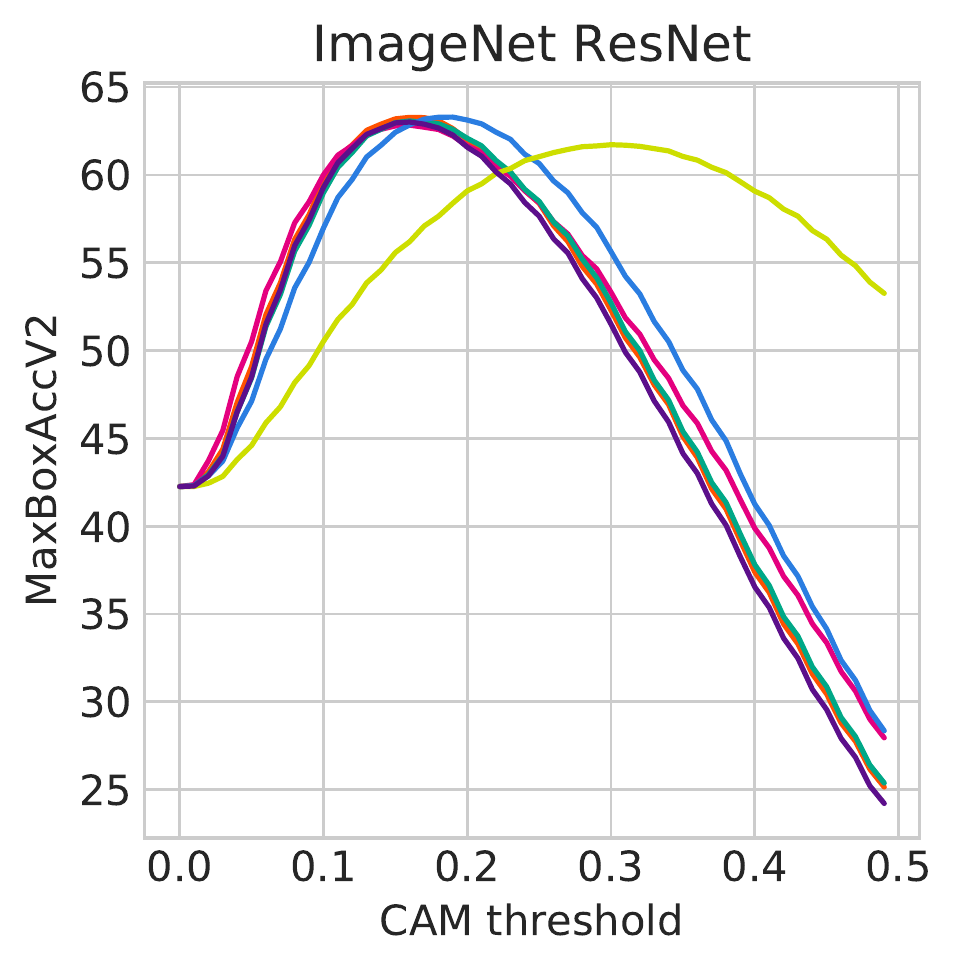}
	\end{subfigure}
	\begin{subfigure}[b]{.31\linewidth}
		\includegraphics[width=\linewidth]{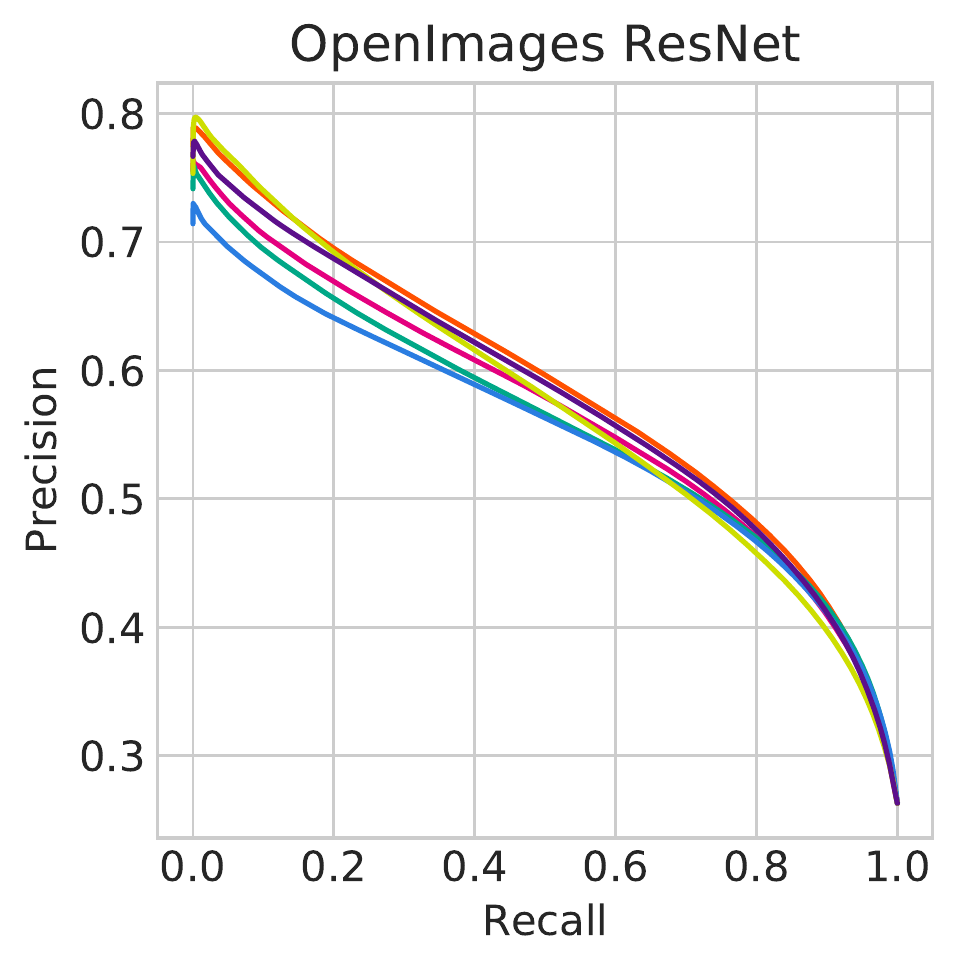}
	\end{subfigure}
	\caption{\small \addition{\textbf{Performance by operating threshold $\tau$.} CUB and ImageNet: \boxacc versus $\tau$, OpenImages: \pxprec versus \pxrec. ResNet architecture results are used.}}
	\label{fig:all_3_by_3_threshold_plots}
\end{figure*}

\begin{figure*}
	\centering
	\begin{subfigure}[b]{0.85\linewidth}
		\begin{subfigure}[b]{.16\linewidth}
			\includegraphics[width=\linewidth]{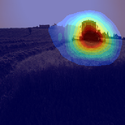}
		\end{subfigure}~\begin{subfigure}[b]{.16\linewidth}
			\includegraphics[width=\linewidth]{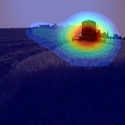}
		\end{subfigure}~\begin{subfigure}[b]{.16\linewidth}
			\includegraphics[width=\linewidth]{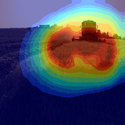}
		\end{subfigure}~\begin{subfigure}[b]{.16\linewidth}
			\includegraphics[width=\linewidth]{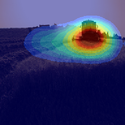}
		\end{subfigure}~\begin{subfigure}[b]{.16\linewidth}
			\includegraphics[width=\linewidth]{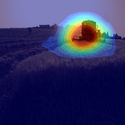}
		\end{subfigure}~\begin{subfigure}[b]{.16\linewidth}
			\includegraphics[width=\linewidth]{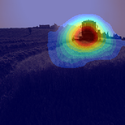}
		\end{subfigure}~\\
		\begin{subfigure}[b]{.16\linewidth}
			\includegraphics[width=\linewidth]{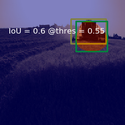}
		\end{subfigure}~\begin{subfigure}[b]{.16\linewidth}
			\includegraphics[width=\linewidth]{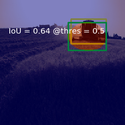}
		\end{subfigure}~\begin{subfigure}[b]{.16\linewidth}
			\includegraphics[width=\linewidth]{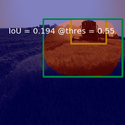}
		\end{subfigure}~\begin{subfigure}[b]{.16\linewidth}
			\includegraphics[width=\linewidth]{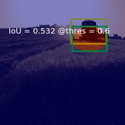}
		\end{subfigure}~\begin{subfigure}[b]{.16\linewidth}
			\includegraphics[width=\linewidth]{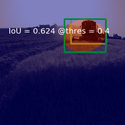}
		\end{subfigure}~\begin{subfigure}[b]{.16\linewidth}
			\includegraphics[width=\linewidth]{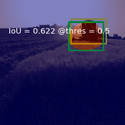}
		\end{subfigure}
	\end{subfigure}
	\caption{\small \addition{\textbf{ImageNet score maps.} Score maps of CAM, HaS, ACoL, SPG, ADL, CutMix from ImageNet.}}
	\label{fig:score_map_visualization_imagenet}
\end{figure*}

\begin{figure*}
	\centering
	\begin{subfigure}[b]{0.85\linewidth}
		\begin{subfigure}[b]{.16\linewidth}
			\includegraphics[width=\linewidth]{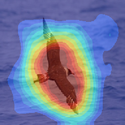}
		\end{subfigure}~\begin{subfigure}[b]{.16\linewidth}
			\includegraphics[width=\linewidth]{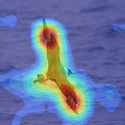}
		\end{subfigure}~\begin{subfigure}[b]{.16\linewidth}
			\includegraphics[width=\linewidth]{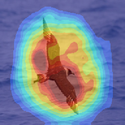}
		\end{subfigure}~\begin{subfigure}[b]{.16\linewidth}
			\includegraphics[width=\linewidth]{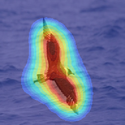}
		\end{subfigure}~\begin{subfigure}[b]{.16\linewidth}
			\includegraphics[width=\linewidth]{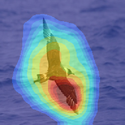}
		\end{subfigure}~\begin{subfigure}[b]{.16\linewidth}
			\includegraphics[width=\linewidth]{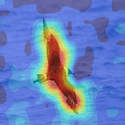}
		\end{subfigure}~\\
		\begin{subfigure}[b]{.16\linewidth}
			\includegraphics[width=\linewidth]{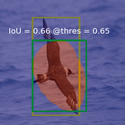}
		\end{subfigure}~\begin{subfigure}[b]{.16\linewidth}
			\includegraphics[width=\linewidth]{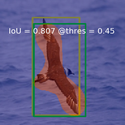}
		\end{subfigure}~\begin{subfigure}[b]{.16\linewidth}
			\includegraphics[width=\linewidth]{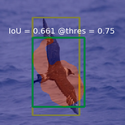}
		\end{subfigure}~\begin{subfigure}[b]{.16\linewidth}
			\includegraphics[width=\linewidth]{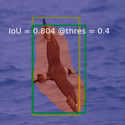}
		\end{subfigure}~\begin{subfigure}[b]{.16\linewidth}
			\includegraphics[width=\linewidth]{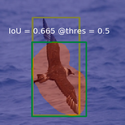}
		\end{subfigure}~\begin{subfigure}[b]{.16\linewidth}
			\includegraphics[width=\linewidth]{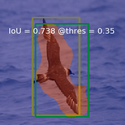}
		\end{subfigure}
	\end{subfigure}
	\caption{\small \addition{\textbf{CUB score maps.} Score maps of CAM, HaS, ACoL, SPG, ADL, CutMix from CUB.}}
	\label{fig:score_map_visualization_cub}
\end{figure*}

\begin{figure*}
	\centering
	\begin{subfigure}[b]{0.85\linewidth}
		\begin{subfigure}[b]{.16\linewidth}
			\includegraphics[width=\linewidth]{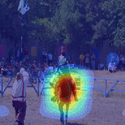}
		\end{subfigure}~\begin{subfigure}[b]{.16\linewidth}
			\includegraphics[width=\linewidth]{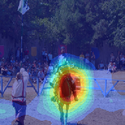}
		\end{subfigure}~\begin{subfigure}[b]{.16\linewidth}
			\includegraphics[width=\linewidth]{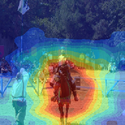}
		\end{subfigure}~\begin{subfigure}[b]{.16\linewidth}
			\includegraphics[width=\linewidth]{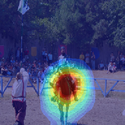}
		\end{subfigure}~\begin{subfigure}[b]{.16\linewidth}
			\includegraphics[width=\linewidth]{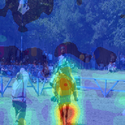}
		\end{subfigure}~\begin{subfigure}[b]{.16\linewidth}
			\includegraphics[width=\linewidth]{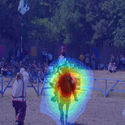}
		\end{subfigure}~\\
		\begin{subfigure}[b]{.16\linewidth}
			\includegraphics[width=\linewidth]{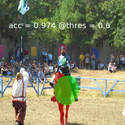}
		\end{subfigure}~\begin{subfigure}[b]{.16\linewidth}
			\includegraphics[width=\linewidth]{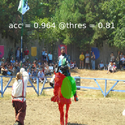}
		\end{subfigure}~\begin{subfigure}[b]{.16\linewidth}
			\includegraphics[width=\linewidth]{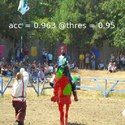}
		\end{subfigure}~\begin{subfigure}[b]{.16\linewidth}
			\includegraphics[width=\linewidth]{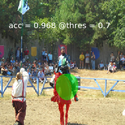}
		\end{subfigure}~\begin{subfigure}[b]{.16\linewidth}
			\includegraphics[width=\linewidth]{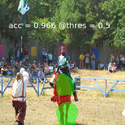}
		\end{subfigure}~\begin{subfigure}[b]{.16\linewidth}
			\includegraphics[width=\linewidth]{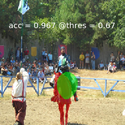}
		\end{subfigure}
	\end{subfigure}
	\caption{\small \addition{\textbf{OpenImages score maps.} Score maps of CAM, HaS, ACoL, SPG, ADL, CutMix from OpenImages.}}
	\label{fig:score_map_visualization_openimages}
\end{figure*}

WSOL evaluation must focus more on score map evaluation, independent of the calibration. As shown in Figure~\ref{fig:qualitative_cam_threshold} the min-max normalized score map for CAM predicts a peaky foreground score on the duck face, While HaS and ACoL score maps show more distributed scores in body areas, demonstrating the effects of adversarial erasing during training. However, the maximal IoU performances do not differ as much. More visual examples are shown in Figures~\ref{fig:score_map_visualization_imagenet},~\ref{fig:score_map_visualization_cub},~and~\ref{fig:score_map_visualization_openimages}.

\begin{figure}
    \centering
    \includegraphics[width=\linewidth]{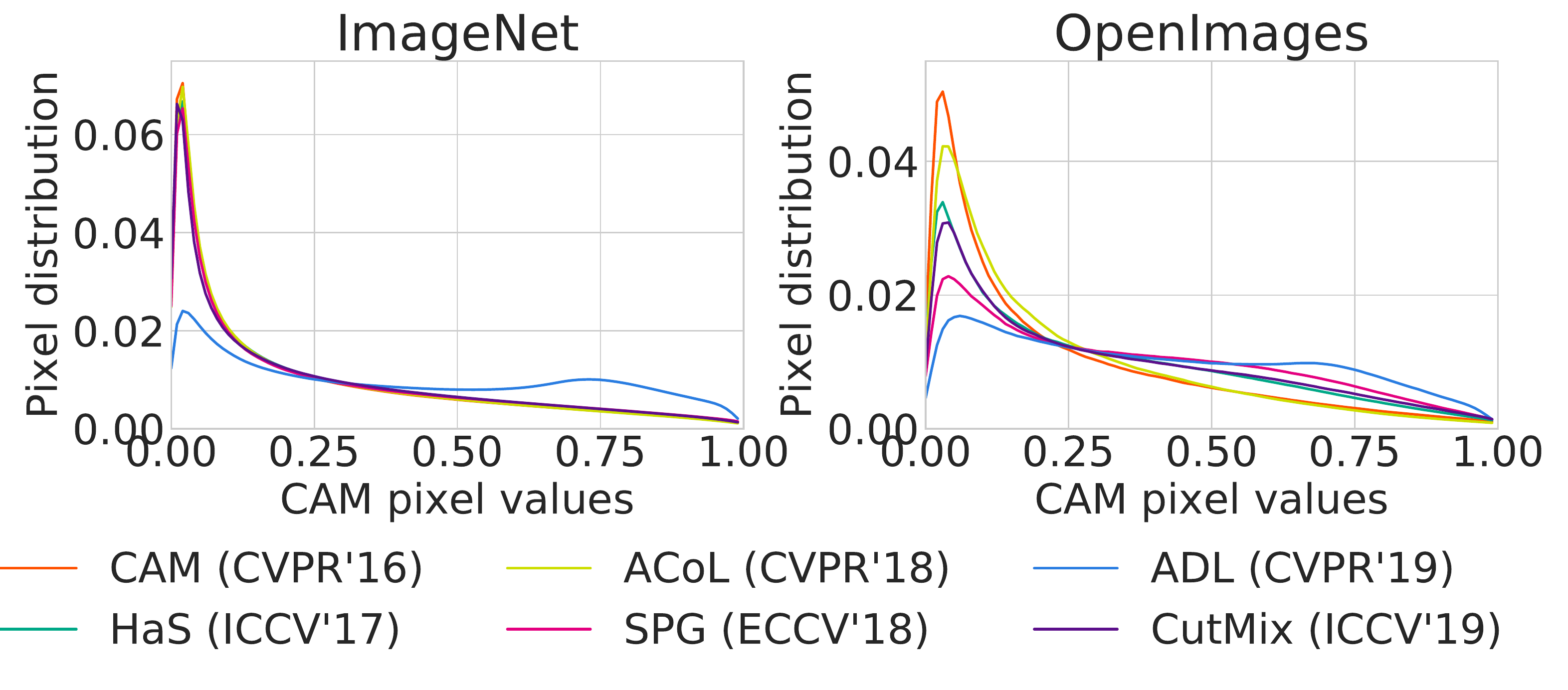}
    \caption{\small \revision{\textbf{CAM pixel value distributions.} On ImageNet and OpenImages \testfullsup.}}
    \label{fig:cam_value_dist}
\end{figure}

This is because WSOL methods exhibit different score distributions. In Figure~\ref{fig:cam_value_dist}, ADL in particular tends to generate flatter score maps. Comparing datasets, we observe that OpenImages tends to have more peaky score distributions. It is therefore important to find the optimal operating point for each method and dataset for fair comparison. 

In Figure~\ref{fig:all_3_by_3_threshold_plots}, we show the performances of the considered methods at different operating thresholds $\tau$. We observe that the optimal operating thresholds $\tau^\star$ are vastly different across data and architectures, while the threshold-independent performances (\newmaxboxacc and \pxap) are not significantly different. Fixing the operating threshold $\tau$ at a pre-defined value, therefore, can lead to an apparent increase in performance without improving the score maps.

\firstround{\myparagraph{Max normalization.} We have mainly used the min-max normalization for post-processing the score maps (Table~\ref{tab:prior_wsol_operating_thresholds}). We show results when max normalization is used instead~\cite{CAM,GradCAM}. Similarly for min-max normalization results in Table~\ref{tab:main_v2}, we show that the methods since CAM do not significantly improve over the CAM baseline. CutMix is the only method that improves over CAM (+0.2\%p); others are generally worse than CAM; in particular, ACoL is significantly worse (-5.4\%p). We thus confirm the same conclusion for max normalization.}

\newcommand\appendixviolinwidth{.33}

\begin{figure*}
	\centering
	
	\begin{subfigure}[b]{\appendixviolinwidth\linewidth}
		\includegraphics[width=\linewidth]{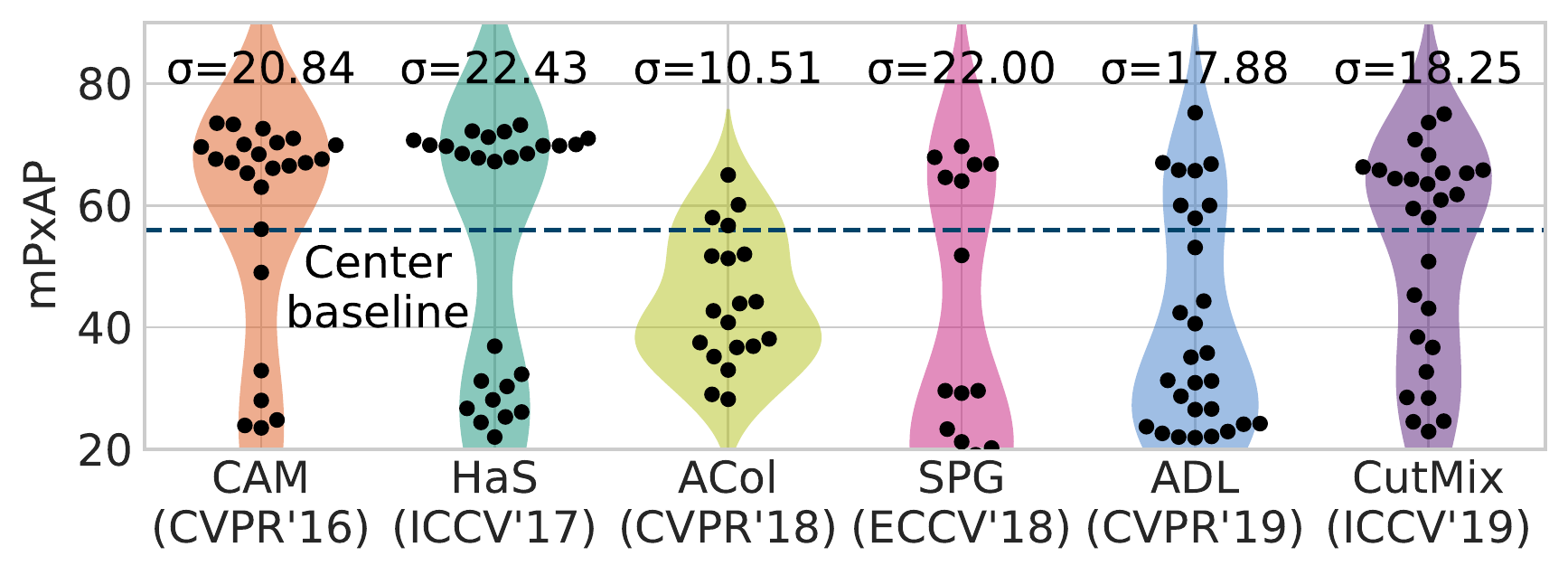}
		\caption{CUB, VGG}
	\end{subfigure}
	\begin{subfigure}[b]{\appendixviolinwidth\linewidth}
		\includegraphics[width=\linewidth]{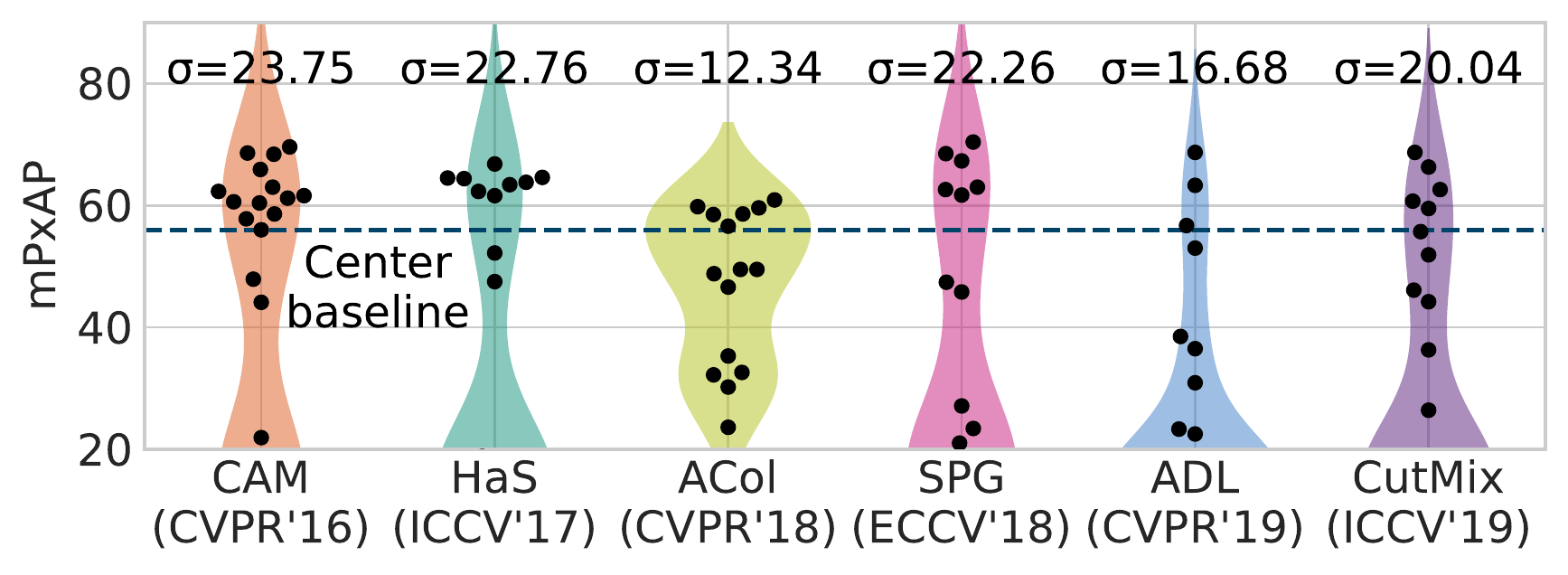}
		\caption{CUB, Inception}
	\end{subfigure}
	\begin{subfigure}[b]{\appendixviolinwidth\linewidth}
		\includegraphics[width=\linewidth]{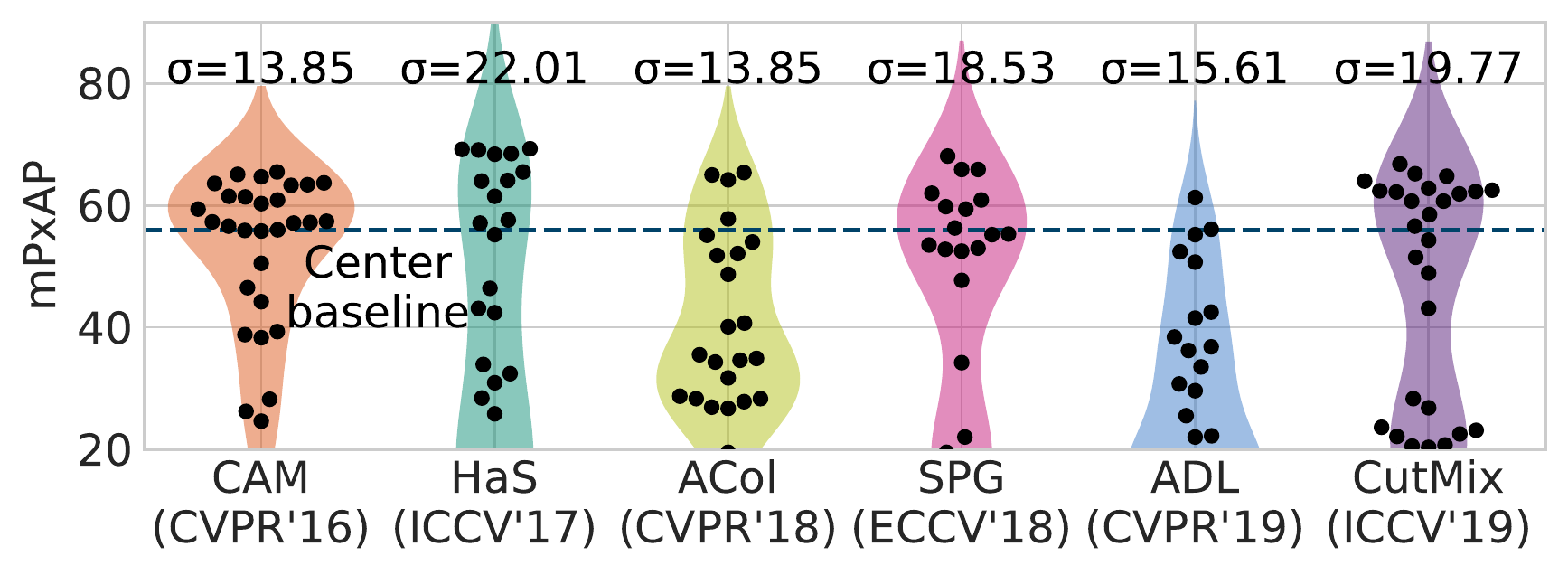}
		\caption{CUB, ResNet}
	\end{subfigure}
	\vspace{1em}
	
	\begin{subfigure}[b]{\appendixviolinwidth\linewidth}
		\includegraphics[width=\linewidth]{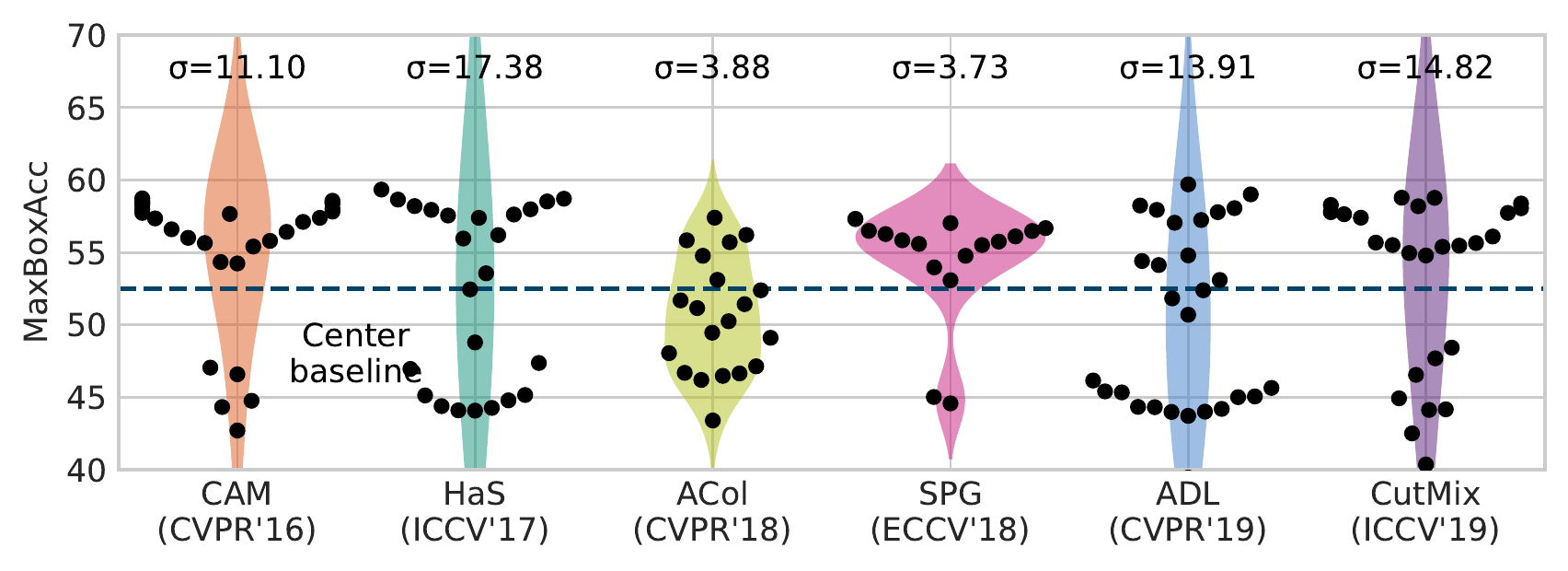}
		\caption{ImageNet, VGG}
	\end{subfigure}
	\begin{subfigure}[b]{\appendixviolinwidth\linewidth}
		\includegraphics[width=\linewidth]{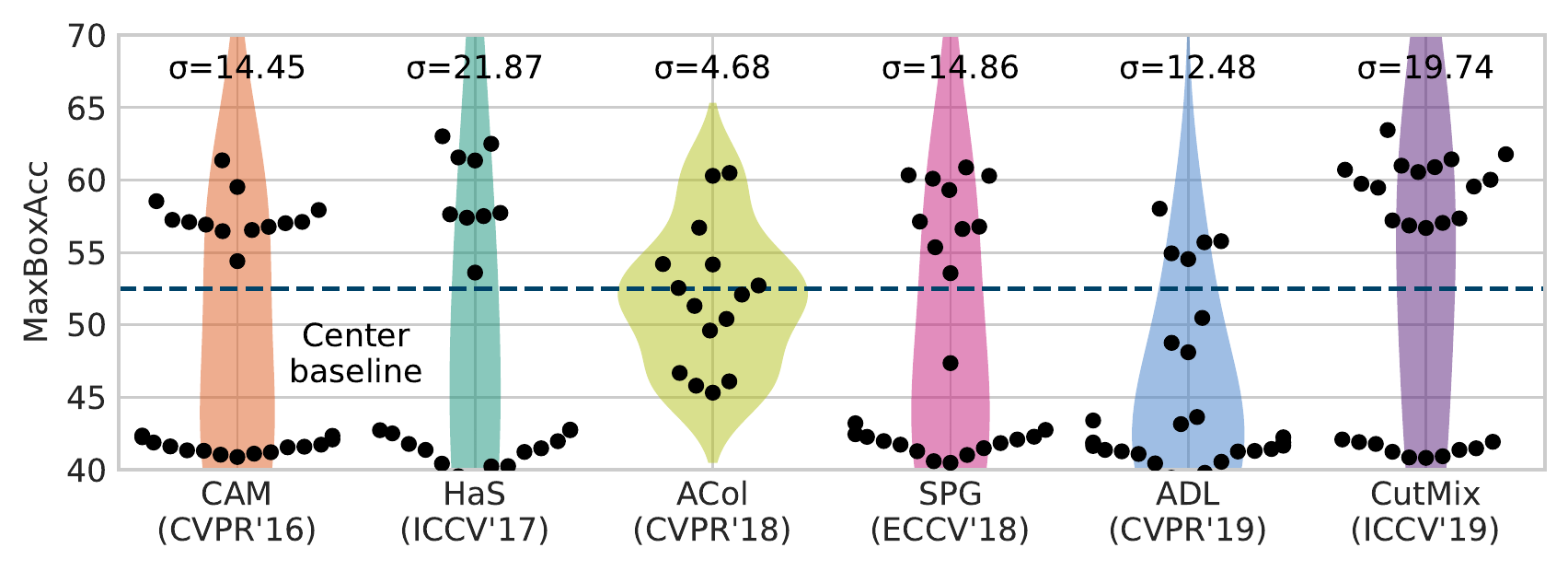}
		\caption{ImageNet, Inception}
	\end{subfigure}
	\begin{subfigure}[b]{\appendixviolinwidth\linewidth}
		\includegraphics[width=\linewidth]{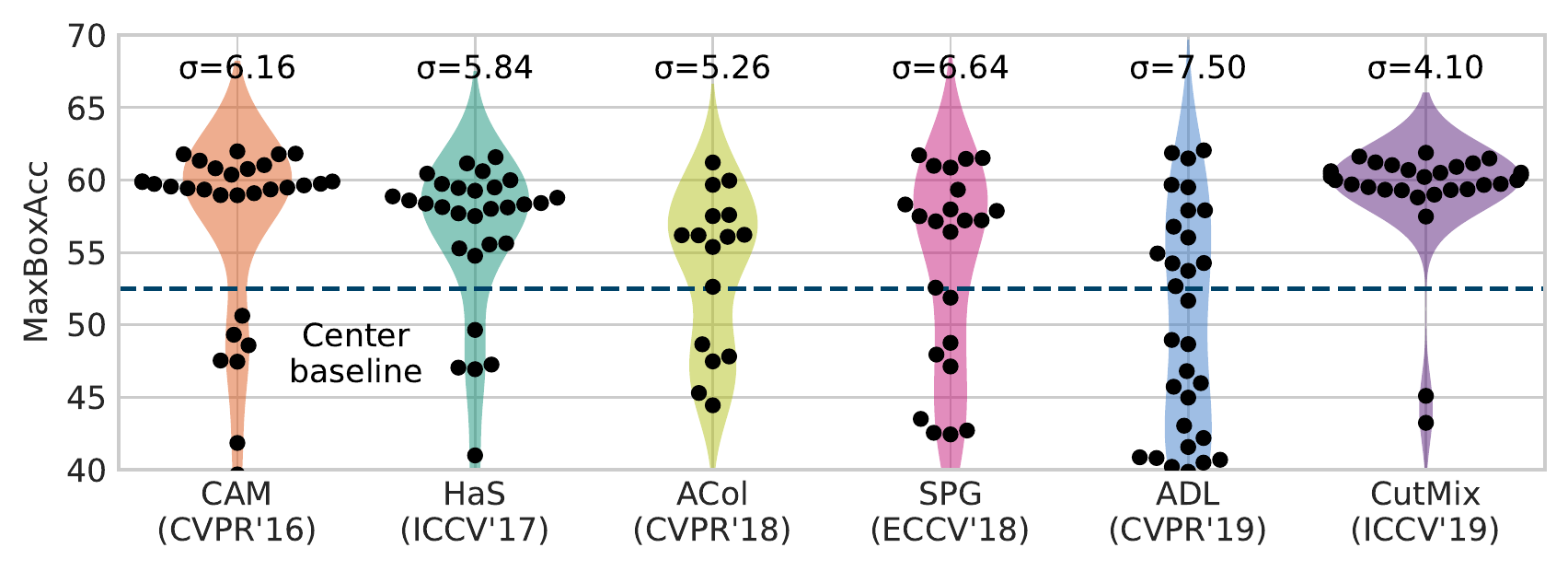}
		\caption{ImageNet, ResNet}
	\end{subfigure}
	\vspace{1em}
	
	\begin{subfigure}[b]{\appendixviolinwidth\linewidth}
		\includegraphics[width=\linewidth]{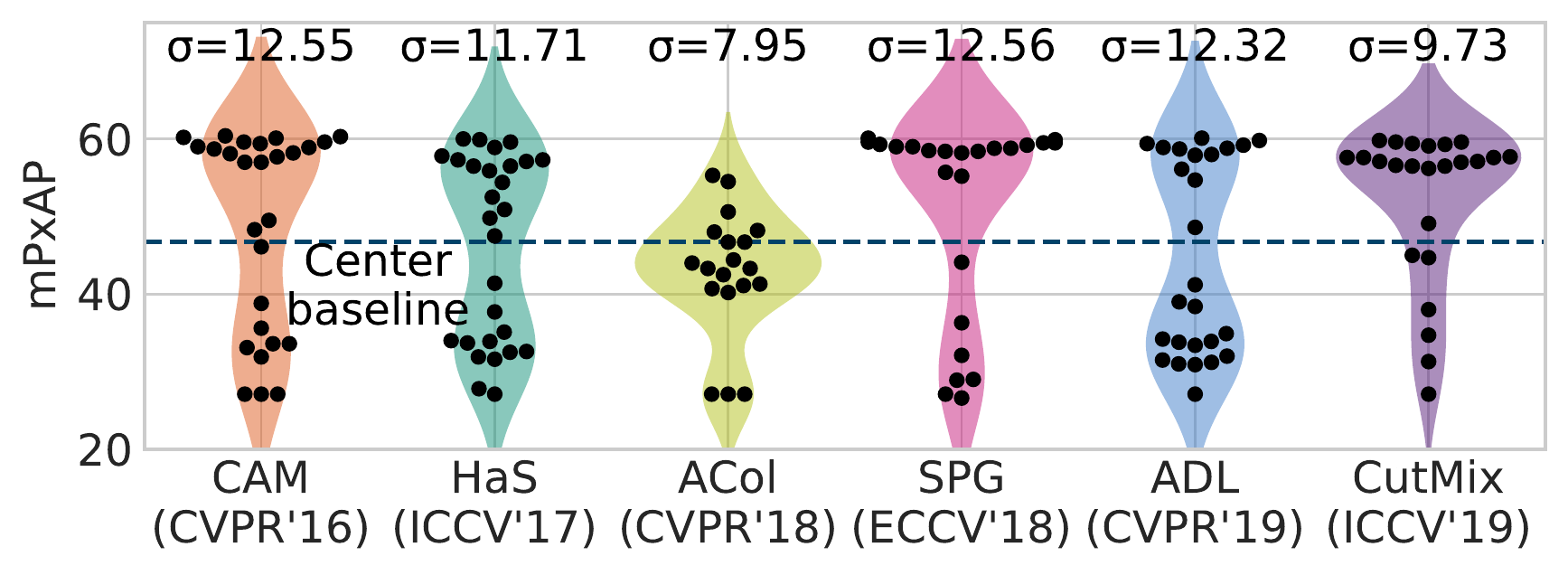}
		\caption{OpenImages, VGG}
	\end{subfigure}
	\begin{subfigure}[b]{\appendixviolinwidth\linewidth}
		\includegraphics[width=\linewidth]{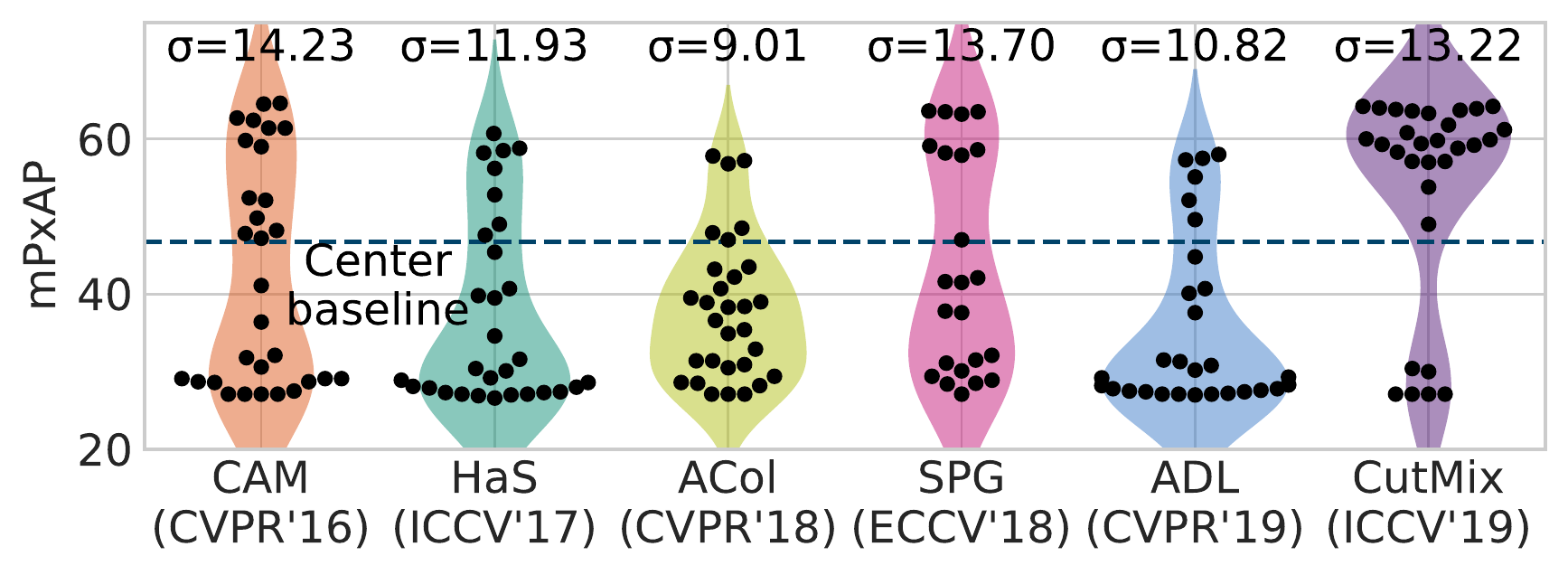}
		\caption{OpenImages, Inception}
	\end{subfigure}
	\begin{subfigure}[b]{\appendixviolinwidth\linewidth}
		\includegraphics[width=\linewidth]{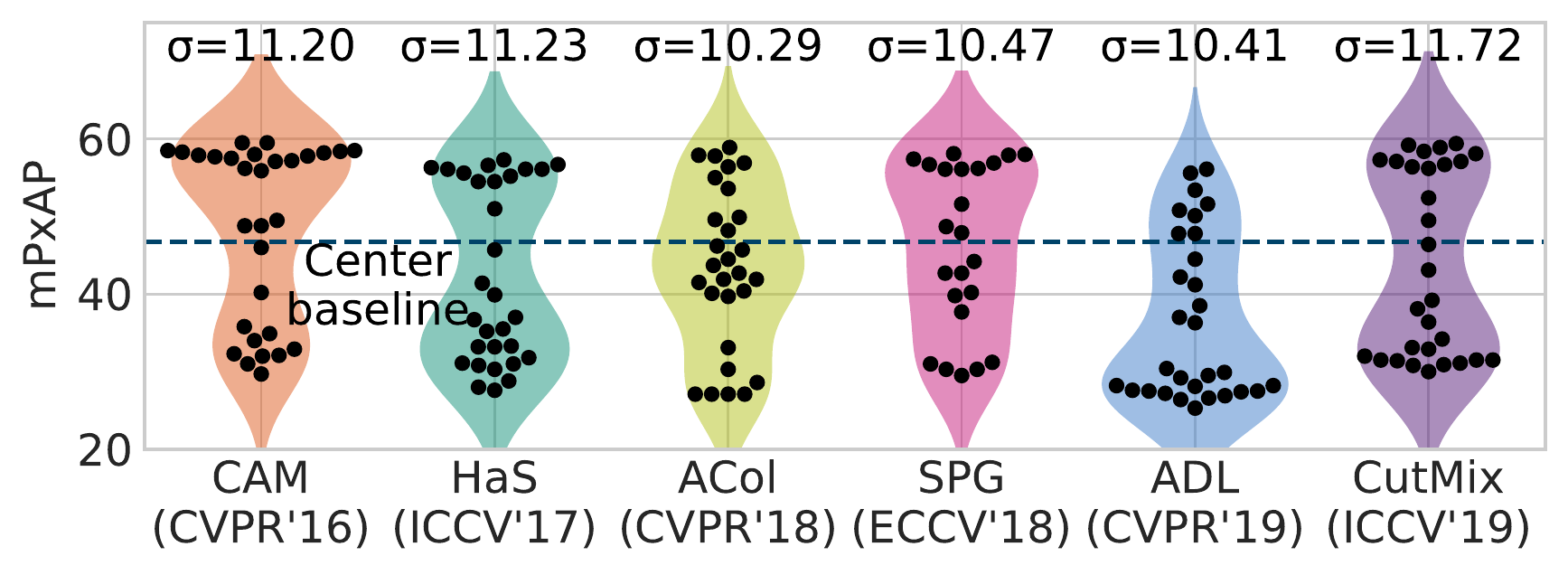}
		\caption{OpenImages, ResNet}
	\end{subfigure}
	\vspace{1em}
	
	\begin{subfigure}[b]{\appendixviolinwidth\linewidth}
		\includegraphics[width=\linewidth]{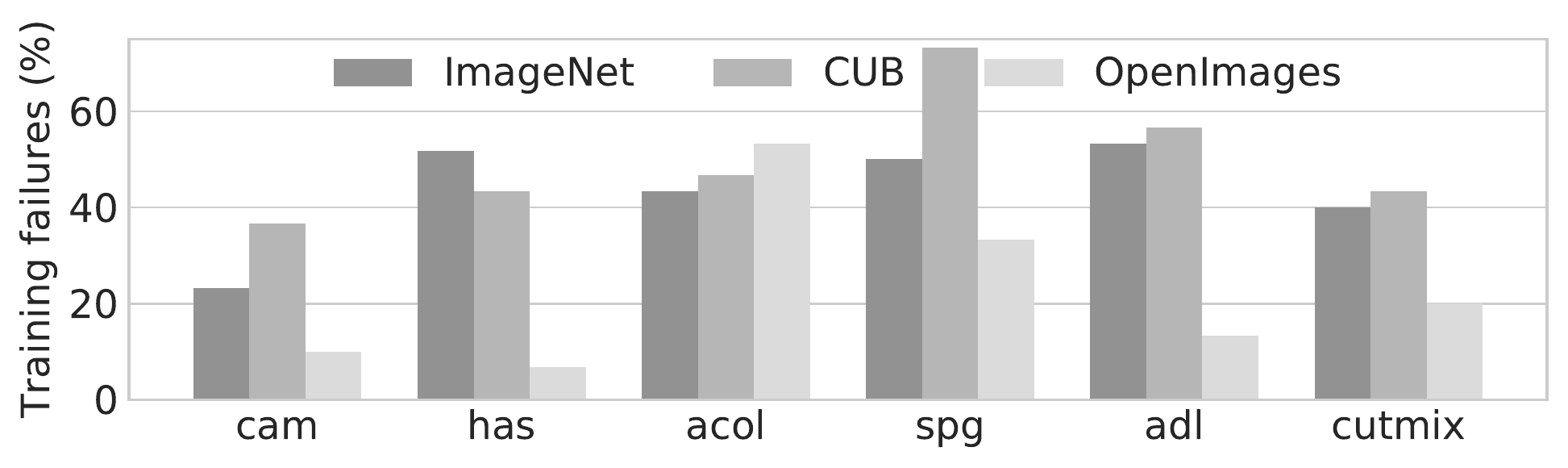}
		\caption{Ratio of training failures, VGG}
		\label{fig:training_failures_vgg}
	\end{subfigure}
	\begin{subfigure}[b]{\appendixviolinwidth\linewidth}
		\includegraphics[width=\linewidth]{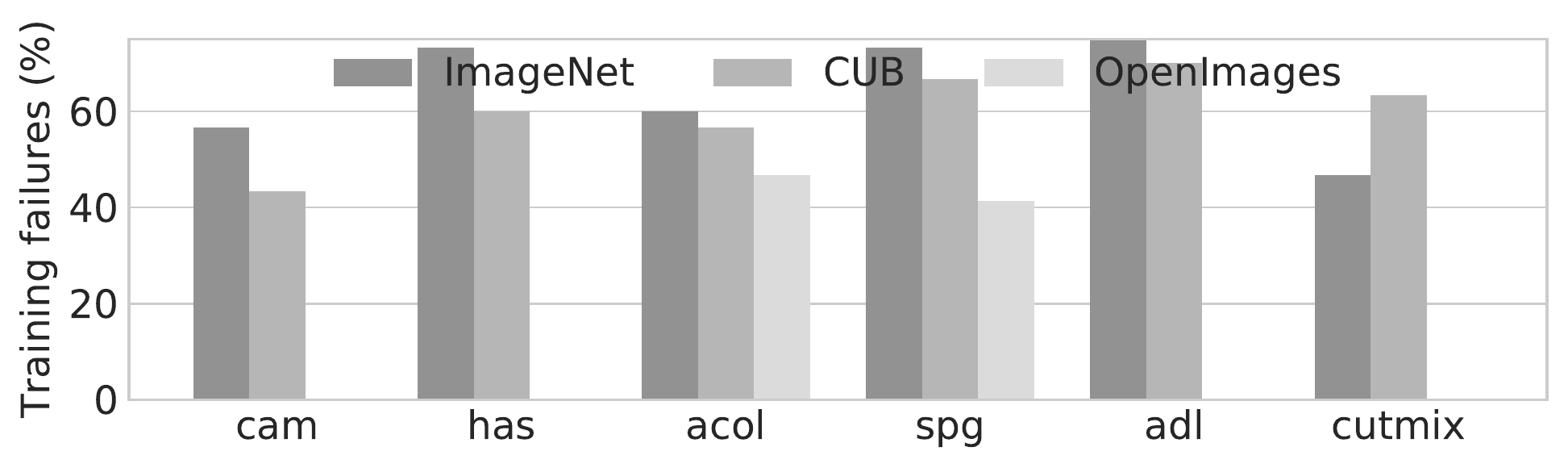}
		\caption{Ratio of training failures, Inception}
		\label{fig:training_failures_inception}
	\end{subfigure}
	\begin{subfigure}[b]{\appendixviolinwidth\linewidth}
		\includegraphics[width=\linewidth]{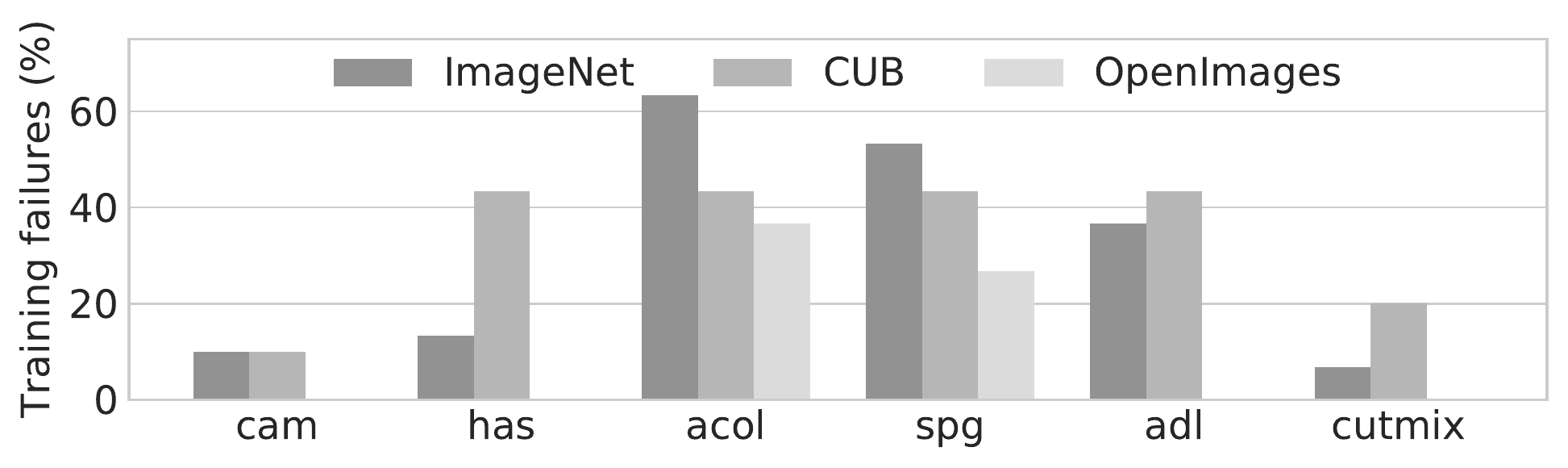}
		\caption{Ratio of training failures, ResNet}
		\label{fig:training_failures_resnet}
	\end{subfigure}
	\vspace{1em}
	
	\caption{\small \addition{\textbf{All results of the 30 hyperparameter trials.} CUB, ImageNet, OpenImages performances of all 30 randomly chosen hyperparameter combinations for each method.}}
	\label{fig:all_3_by_3_violin_plots}
\end{figure*}

\subsection{Hyperparameter analysis}
\label{subsec:hyperparameter_analysis}

Different types and amounts of full supervision used in WSOL methods manifest in the form of model hyperparameter selection (\S\ref{sec:wsol_impossibility}). Here, we measure the impact of the validation on \trainfullsup by observing the performance distribution among 30 trials of random hyperparameters. We then study the effects of feature-erasing hyperparameters, a common hyperparameter type in WSOL methods.

\begin{figure}[t]
	\centering
	\definecolor{greencross}{RGB}{60,140,50}
	\centering
	\includegraphics[width=0.34\linewidth]{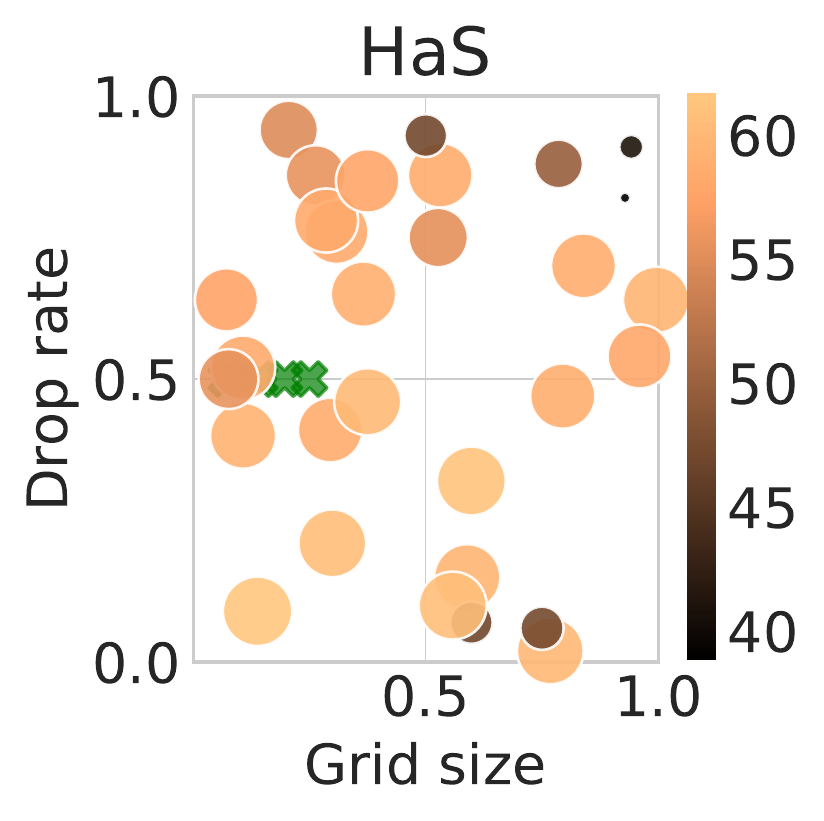}\hspace{0.03\linewidth}%
	\includegraphics[width=0.22\linewidth]{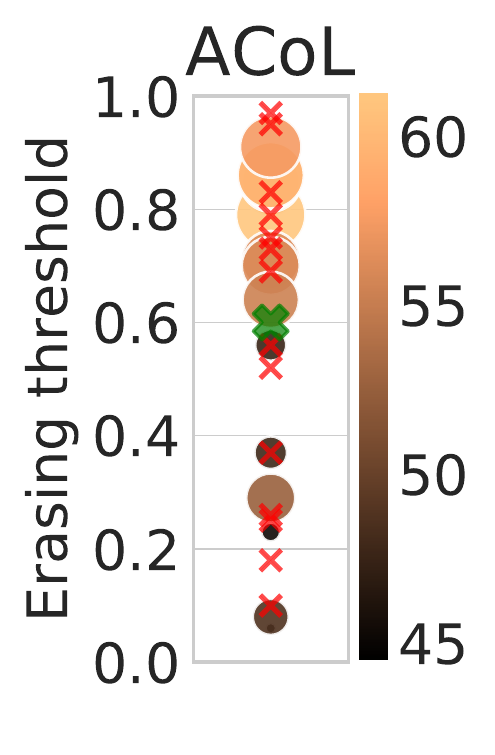}\hspace{0.03\linewidth}%
	\includegraphics[width=0.34\linewidth]{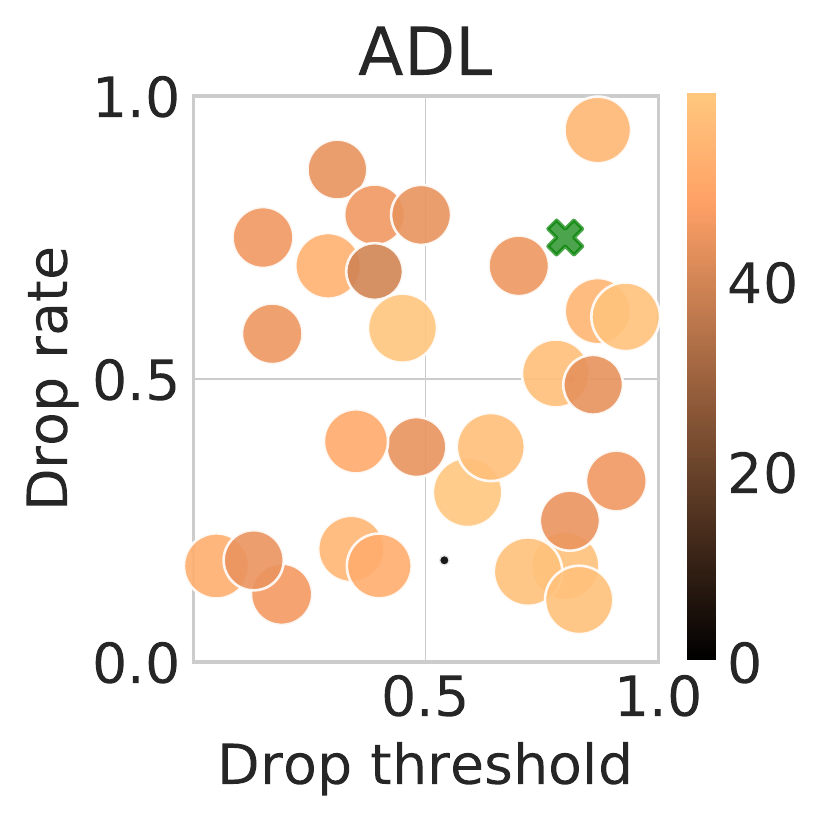}
	\caption{\small \revision{\textbf{Impact of hyperparameters for feature erasing.} Color and size of the circles indicate the performance at the corresponding hyperparameters. {\color{red}\ding{53}}: non-convergent training sessions. {\color{greencross}\ding{54}}: hyperparameters suggested by the original papers.}
	}
	\label{fig:adl_acol_scatter}
	\vspace{-1.1em}
\end{figure}

\myparagraph{Performance with 30 hyperparameter trials.}
To measure the sensitivity of each method to hyperparameter choices, we plot the performance distribution of the intermediate models in the 30 random search trials. \revision{We say that a training session is \textit{non-convergent} if the training loss is \texttt{nan} at the last epoch.} We show the performance distributions of the converged sessions, and report the ratio of non-convergent sessions separately.

\revision{Our results in Figure~\ref{fig:all_3_by_3_violin_plots} indicate the diverse range of performances depending on the hyperparameter choice. Specifically, we observe that (1) Performances do vary according to the hyperparameter choice, so the hyperparameter optimization is necessary for the optimal performances. (2) CAM is among the more stable WSOL methods. (3) ACoL and ADL show greater sensitivity to hyperparameters in general. (4) CUB is a difficult benchmark where random choice of hyperparameters is highly likely to lead to performances worse than the center-Gaussian baseline. We thus suggest to use the vanilla CAM when absolutely no full supervision is available.}

Figure~\ref{fig:all_3_by_3_violin_plots} (a-c) shows that WSOL on CUB are generally struggling: random hyperparameters often show worse performance than the center baseline. We conjecture that CUB is a disadvantageous setup for WSOL: as all images contain birds, the models only attend on bird parts for making predictions. We believe adding more non-bird images can improve the overall performances~(\S\ref{subsec:when_is_wsol_unsolvable}). 

We show the non-convergence statistics in Figure~\ref{fig:all_3_by_3_violin_plots} (j-l). Vanilla CAM exhibit a stable training: non-convergence rates are low on all three datasets. ACoL, SPG, and ADL suffer from many training failures, especially on CUB.

In conclusion, vanilla CAM is stable and robust to hyperparameters.
Complicated design choices introduced by later methods only seem to lower the overall performances rather than providing new avenues for performance boost.

\begin{figure*}
	\centering
	\begin{subfigure}[b]{.30\linewidth}
		\includegraphics[width=\linewidth]{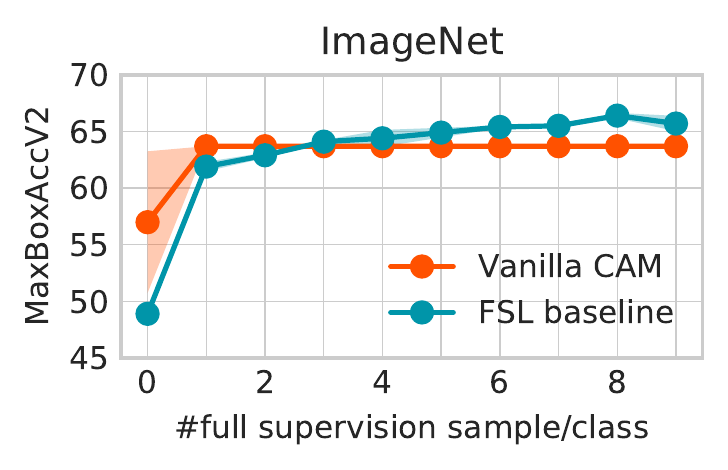}
	\end{subfigure}
	\begin{subfigure}[b]{.30\linewidth}
		\includegraphics[width=\linewidth]{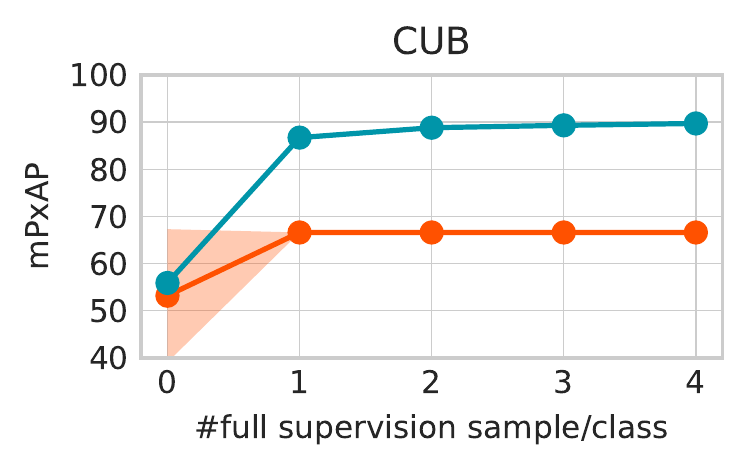}
	\end{subfigure}
	\begin{subfigure}[b]{.30\linewidth}
		\includegraphics[width=\linewidth]{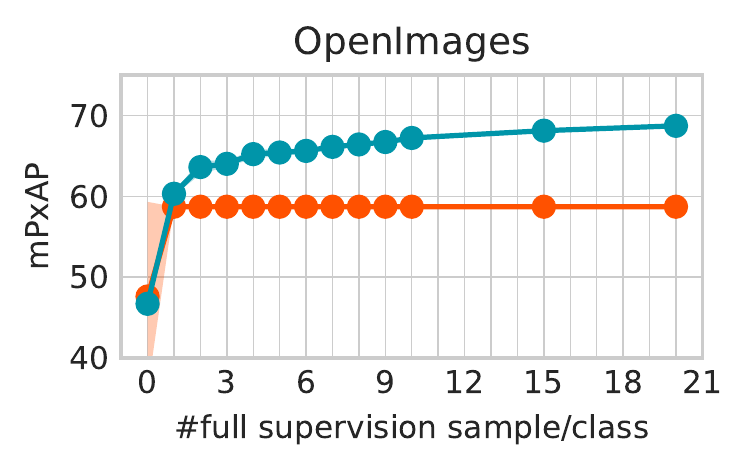}
	\end{subfigure}
	\caption{\small \revision{\textbf{WSOL versus few-shot learning.} The mean and standard error of models trained on three samples of full-supervision subsets are reported. ResNet50 is used throughout. At 0 full supervision, Vanilla CAM$=$random-hyperparameter and FSL$=$center-gaussian baseline.}
	}
	\label{fig:wsol_vs_fsl}
\end{figure*}

\myparagraph{Effects of erasing hyperparameters.}
Many WSOL methods since CAM have introduced different forms of erasing to encourage models to extract cues from broader regions (\S\ref{subsec:prior_wsol_methods}). We study the contribution of such hyperparameters in ADL, HaS, and ACoL in Figure~\ref{fig:adl_acol_scatter}. We observe that the performance improves with higher erasing thresholds (ADL drop threshold and ACoL erasing threshold). We also observe that lower drop rates leads to better performances (ADL and HaS). The erasing hyperparameters introduced since CAM only negatively impact the performance.

\begin{figure*}
	\centering
		\begin{subfigure}[b]{.25\linewidth}
		\includegraphics[width=\linewidth, page=1]{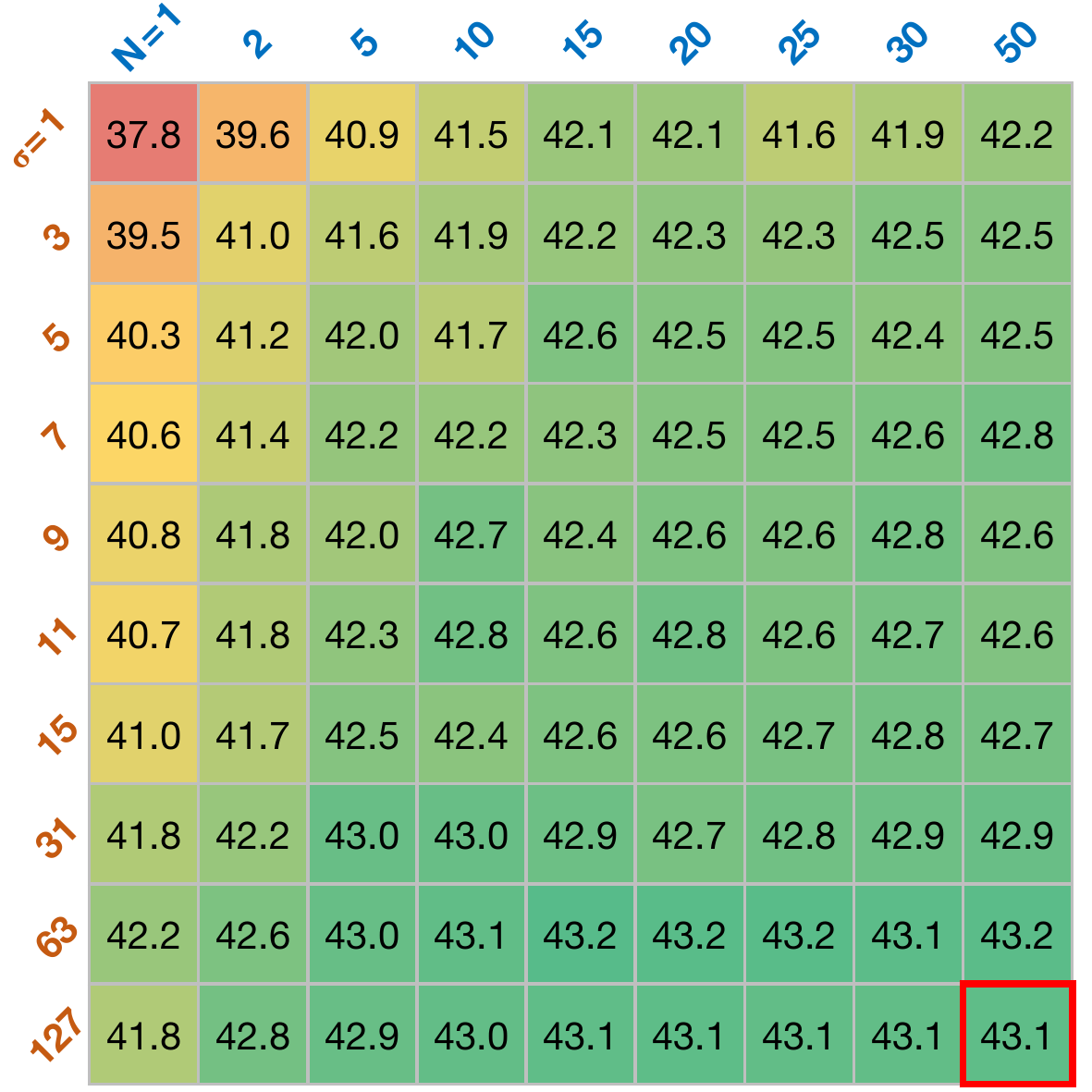}
		\caption{SG-VB}
	\end{subfigure}%
	\begin{subfigure}[b]{.25\linewidth}
		\includegraphics[width=\linewidth, page=2]{figures/saliency_hyperparams.pdf}
		\caption{SG-GB}
	\end{subfigure}%
	\begin{subfigure}[b]{.25\linewidth}
		\includegraphics[width=\linewidth, page=3]{figures/saliency_hyperparams.pdf}
		\caption{IG-VB}
	\end{subfigure}%
	\begin{subfigure}[b]{.25\linewidth}
		\includegraphics[width=\linewidth, page=4]{figures/saliency_hyperparams.pdf}
		\caption{IG-GB}
	\end{subfigure}%
	\caption{\small \addition{\textbf{Hyperparameter selection for saliency methods.} Each cell shows OpenImages 30k \pxap for the combinations of SmoothGrad (SG), integrated gradient (IG), vanilla backpropagation (VB), and guided backpropagation (GB). The red bordered cells denote the hyperparameter combination used for latter experiments.}}
	\label{fig:saliency_hyperparameters}
\end{figure*}

\addition{
\subsection{Visual interpretability methods as WSOL methods}
\label{subsec:saliency_methods}

Visual interpretability methods have appeared in the community as a branch relatively independent of the CAM~\cite{CAM} variants considered above. They are designed to shed light on the reasoning behind the decisions made by learned models. For image classifiers, the most popular form of visual interpretation method is \textit{input attribution}. Given an input image and a model, an input attribution method produces a score map indicating the contribution of each pixel towards the model decision.

While visual interpretability methods are often evaluated with the dedicated tests for explainability (see \cite{adebayo2018sanity, hooker2019benchmark} for an overview), we observe that they are eventually algorithms for producing score maps indicating the pixels that are likely to contain the cues for recognition. In this interlude section, we examine the potential of visual interpretability methods for tackling the WSOL problem. The CAM paper~\cite{CAM} has included the baseline results for input gradients, the most basic input attribution method, but no work since then has systematically evaluated the interpretability methods in terms of the localization performance.

\myparagraph{Visual interpretability methods.}
We evaluate four variants in this study.
\textbf{Vanilla backpropagation~(VB)~\cite{FirstDNNInputGradient}}  computes the input-gradient score map. It measures the local contribution of each pixel towards the model outputs.
\textbf{Guided backpropagation~(GB)~\cite{GuidedBackprop}} is a modified version of backpropagation for DNNs with ReLU activations. Unlike VB, GB also applies the ReLU activation during the backward pass.
\textbf{SmoothGrad~(SG)~\cite{SmoothGrad}} is designed to overcome the limitation of VB that it only considers the model responses to infinitesimal changes around the input RGB values. SG averages the input gradients for multiple noised versions of the input image. The number of noised versions of the input, $N_{\mathrm{SG}}$, is a hyperparameter.
\textbf{Integrated gradient~(IG)~\cite{IntegratedGradients}} is another method that addresses the locality limitation of VB. IG averages the input gradients along the interpolated images from the reference image to a zero (black) image. For this method, the number of data points $N_{\mathrm{IG}}$ is a hyperparameter.
Note that the choice of (pseudo-)input gradient generation algorithm (VB or GB) and the choice of input synthesis and aggregation algorithm (SG, IG, or None) are orthogonal. We thus experiment with all six possible combinations: VB, GB, SG-VB, SG-GB, IG-VB, and IG-GB. Interestingly, we observe that the localization performance does not improve for IG-GB while $N_{IG}$ increases. We conjecture that this is because GB highlights only the edge of the image, rather than interprets the model decision~\cite{adebayo2018sanity}. More specifically, the same score maps are produced from the dimmed images, so the localization performance remains the same.

\myparagraph{Processing score maps.}
Unlike CAM score maps, input gradient variants tend to be noisy and peaky. We consider the option to smooth out the score maps via Gaussian blurring, as done in \cite{oh2017exploiting}. The kernel size $\sigma$ is a hyperparameter.

\myparagraph{Which hyperparameters to use?}
We conduct preliminary experiments to decide the hyperparameters $\sigma$, $N_{IG}$, and $N_{SG}$. Figure~\ref{fig:saliency_hyperparameters} summarize the results on the \trainfullsup split. We observe that large $\sigma$ and $N$ improve object localization performance for both IG and SG, and the performance gain saturates when $\sigma$ and $N$ are sufficiently large. We use $N_{\text{IG}} = N_{\text{SG}} = 50$ and $\sigma = 127$ in this paper.

\setlength{\tabcolsep}{3pt}
\begin{table}
\centering
\small
\firstround{\begin{tabular}{lc*{3}{c}g}
		Methods  &  & VGG & Inception & ResNet & Mean \\
		\cline{1-1}\cline{3-6} & \vspace{-.75em}\\
		VB~\cite{FirstDNNInputGradient} &  & 55.6 & 55.1 & 53.7 & 54.8 \\
		GB~\cite{GuidedBackprop} &  & 55.2 & 54.6 & 52.7 & 54.2 \\
		SG-VB~\cite{SmoothGrad, FirstDNNInputGradient} &  & 49.1 & 48.3 & 39.6 & 44.4 \\
		SG-GB~\cite{SmoothGrad, GuidedBackprop} &  & 43.2 & 47.8 & 43.7 & 44.9 \\
		IG-VB~\cite{IntegratedGradients, FirstDNNInputGradient} &  & 58.6 & 57.3 & 56.7 & 57.5 \\
		IG-GB~\cite{IntegratedGradients, GuidedBackprop} &  & 54.5 & 57.5 & 51.7 & 54.6 \\
		\cline{1-1}\cline{3-6} & \vspace{-1em} \\
		CAM~\cite{CAM} &  & \textbf{59.2} & \textbf{63.6} & \textbf{58.7} & \textbf{60.5} \\
		\cline{1-1}\cline{3-6} & \vspace{-.75em} \\
\end{tabular}}
\caption{\small \firstround{\textbf{WSOL evaluation for visual interpretability methods.} \mpxap performances on the OpenImages \testfullsup split.}}
\label{tab:saliency_result}
\end{table}

\myparagraph{Evaluation setup.}
We evaluate the score maps from the above input attribution methods on OpenImages30k using our WSOL evaluation framework. Following our hyperparameter search protocol (\S\ref{subsec:hyperparameter_analysis}), we randomly sample 30 training hyperparameters and select the best hyperparameter combination based on the \trainfullsup performance. Note that we fix the hyperparameters $\sigma$ and $N$ because they are hyperparameters for inference. In addition, we use the same checkpoints used for evaluating CAM to factor out the influence of training process.

\myparagraph{Results.}
Table~\ref{tab:saliency_result} summarize the experimental results. We observe that there is a meaningful ranking among the input attribution methods. The VB variants are mostly better than the GB variants: average \pxap scores across architectures are (54.0, 49.6, 57.1) for VB variants and (53.4, 43.8, 53.9) for GB variants (in the order of none, SG, and IG), respectively.
We observe that SG decreases the localization performance of VB (54.0 to 49.6 architecture-mean \pxap) and GB (53.4 to 43.8 architecture-mean \pxap). On the other hand, IG significantly improves the performance of VB (54.0 to 57.1 architecture-mean \pxap) and GB (53.4 to 53.9 architecture-mean \pxap). Yet, the overall performance of input attribution methods falls behind the CAM baseline. Even the best input attribution performance (58.2 by IG-VB with VGG) is upper bounded by the worst performance of CAM (58.3 with VGG).

\myparagraph{Conclusion.}
We conclude the interlude with the following observations. (1) Vanilla backpropagation (VB) is better than the guided backpropagation (GB) for object localization. (2) SmoothGrad (SG) is not an effective synthesis and aggregation strategy. (3) IntegratedGradients (IG) improves localization. (4) CAM performs better than all of the considered input attribution techniques.
}

\subsection{Few-shot learning baselines}
\label{subsec:few_shot_learning_results}

\revision{
We have discussed the conceptual need for localization supervision (\S\ref{sec:wsol_impossibility}) and the corresponding experimental results where the localization supervision \trainfullsup is used for searching hyperparameters in prior WSOL methods (\S\ref{subsec:main_comparison_wsol}). Given \trainfullsup, one may then use the localization supervision for training the model itself, rather than for finding hyperparameters. We investigate the model performances when under this few-shot learning (FSL) paradigm. The architecture and loss for the FSL baseline models are introduced in \S\ref{subsec:prior_wsol_methods}.
}

\addition{
In the conference version~\cite{wsoleval}, the FSL models have used 100\% of the \trainfullsup split for model training. Since the FSL models also have their own set of hyperparameters, it is not realistic to set them without validation. In this journal version, we perform a validation with 20\% of \trainfullsup to search the FSL hyperparameters (learning rate and feature map size). Then, the found hyperparameters are used for learning the final model with 100\% of \trainfullsup. The performance reports are based on the \testfullsup, as for the WSOL experiments in \S\ref{subsec:main_comparison_wsol}.
}

Performances of the FSL baselines are presented in Table~\ref{tab:main_v2}. Our simple FSL method performs better than the vanilla CAM at $10$, $5$, and $25$ fully labeled samples per class for ImageNet, CUB, and OpenImages, respectively. \revision{The mean FSL performances on CUB and OpenImages are 86.1\% and 75.2\%, which is far better than those of the maximal WSOL performance of 61.1\% and 60.0\%. The results suggests that the FSL baseline is a strong baseline to beat.} 

We compare FSL against CAM at different sizes of \trainfullsup in Figure~\ref{fig:wsol_vs_fsl}.
We simulate the zero-fully-labeled WSOL performance with a set of randomly chosen hyperparameters (\S\ref{subsec:hyperparameter_analysis}); for FSL, we simulate the no-learning performance through the center-Gaussian baseline.

\revision{FSL baselines surpass the CAM results already at 1 full supervision per class for CUB and OpenImages (80.9 and 68.2\% \maxboxacc and \pxap).} We attribute the high FSL performance on CUB to the fact that all images are birds; with 1 sample/class, there are effectively 200 birds as training samples. For OpenImages, the high FSL performance is due to the rich supervision provided by pixel-wise masks. \addition{Interestingly, the performance of WSOL is worse than that of center baseline on CUB (45.0\% and 54.4\% \newmaxboxacc). We believe that this is because most birds are located on the center of images.} \revision{On ImageNet, FSL results are not as great: they surpass the CAM result at 3 samples per class (64.1\%).} Overall, however, FSL performances are strikingly good, even at a low data regime. 
Thus, given a few fully labeled samples, it is perhaps better to train a model with them than to search hyperparameters.

\section{Discussion and Conclusion}
\label{sec:conclusion}

After years of weakly-supervised object localization (WSOL) research, we look back on the common practice and make a critical appraisal. Based on a precise definition of the task, we have argued that WSOL is ill-posed and have discussed how previous methods have used different types of implicit full supervision (\eg tuning hyperparameters with pixel-level annotations) to bypass this issue (\S\ref{sec:wsol_impossibility}). We have then proposed an improved evaluation protocol that allows the hyperparameter search over a few labeled samples (\S\ref{sec:evaluation}). Our empirical studies lead to some striking conclusions: CAM is still not worse than the follow-up methods (\S\ref{subsec:main_comparison_wsol}) and it is perhaps better to use the full supervision directly for model fitting, rather than for hyperparameter search (\S\ref{subsec:few_shot_learning_results}).

We propose the following future research directions for the field. (1) Resolve the ill-posedness via \eg adding more background-class images (\S\ref{subsec:when_is_wsol_unsolvable}). (2) Define the new task, \textit{semi-weakly-supervised object localization}, where methods incorporating both weak and full supervision are studied. 

Our work has implications in other tasks where learners are not supposed to be given full supervision, but are supervised implicitly via model selection and hyperparameter fitting. Examples include weakly-supervised vision tasks (\eg detection and segmentation), zero-shot learning, and unsupervised tasks (\eg disentanglement~\cite{Disentangle}).

{
\myparagraph{Acknowledgements.}
We thank Dongyoon Han, Hyojin Park, Jaejun Yoo, Jung-Woo Ha, Junho Cho, Kyungjune Baek, Muhammad Ferjad Naeem, Rodrigo Benenson, Youngjoon Yoo, and Youngjung Uh for the feedback. NAVER Smart Machine Learning (NSML)~\cite{NSML} has been used. Graphic design by Kay Choi. 
The work is supported by the Basic Science Research Program through NRF funded by the MSIP (NRF-2019R1A2C2006123, 2020R1A4A1016619)  the IITP grant funded by the MSIT (2020-0-01361, Artificial Intelligence Graduate School Program (YONSEI UNIVERSITY)), and the Korea Medical Device Development Fund grant funded by the Korean government (Project Number:  202011D06). This work was also funded by DFG-EXC-Nummer 2064/1-Projektnummer 390727645 and the ERC under the Horizon 2020 program (grant agreement No. 853489).
}%

{\small
\bibliographystyle{IEEEtran}
\bibliography{ref}
}

\begin{IEEEbiography}[{\includegraphics[width=1in,height=1.15in,clip,keepaspectratio]{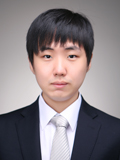}}]{Junsuk Choe} is an assistant professor in the Department of Computer Science and Engineering at Sogang University. Before that, he was a research scientist at NAVER AI Lab (2020-21). He received his Ph.D degree in the School of Integrated Technology at Yonsei University (2020). He obtained his B.S. degree from the School of Electrical and Electronic Engineering at Yonsei University (2013). His recent research interests are in computer vision and machine learning. In particular, he is interested in learning from imperfect data, such as weakly-supervised learning and semi-supervised learning.
\end{IEEEbiography}

\begin{IEEEbiography}[{\includegraphics[width=1in,height=1.15in,clip,keepaspectratio]{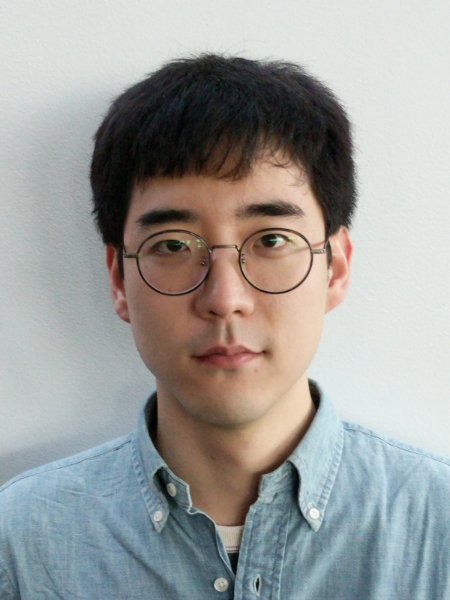}}]{Seong Joon Oh} is a research scientist at NAVER AI Lab. He received his PhD in Machine Learning and Computer Vision under the supervision of Prof. Dr. Bernt Schiele and Prof. Dr. Mario Fritz from the Max Planck Institute for Informatics in 2018, with the thesis "Image manipulation against learned models: privacy and security implications". He received his Master's and Bachelor's degree in Mathematics from the University of Cambridge in 2014. His research interests are in computer vision and machine learning.
\end{IEEEbiography}

\begin{IEEEbiography}[{\includegraphics[width=1in,height=1.15in,clip,keepaspectratio]{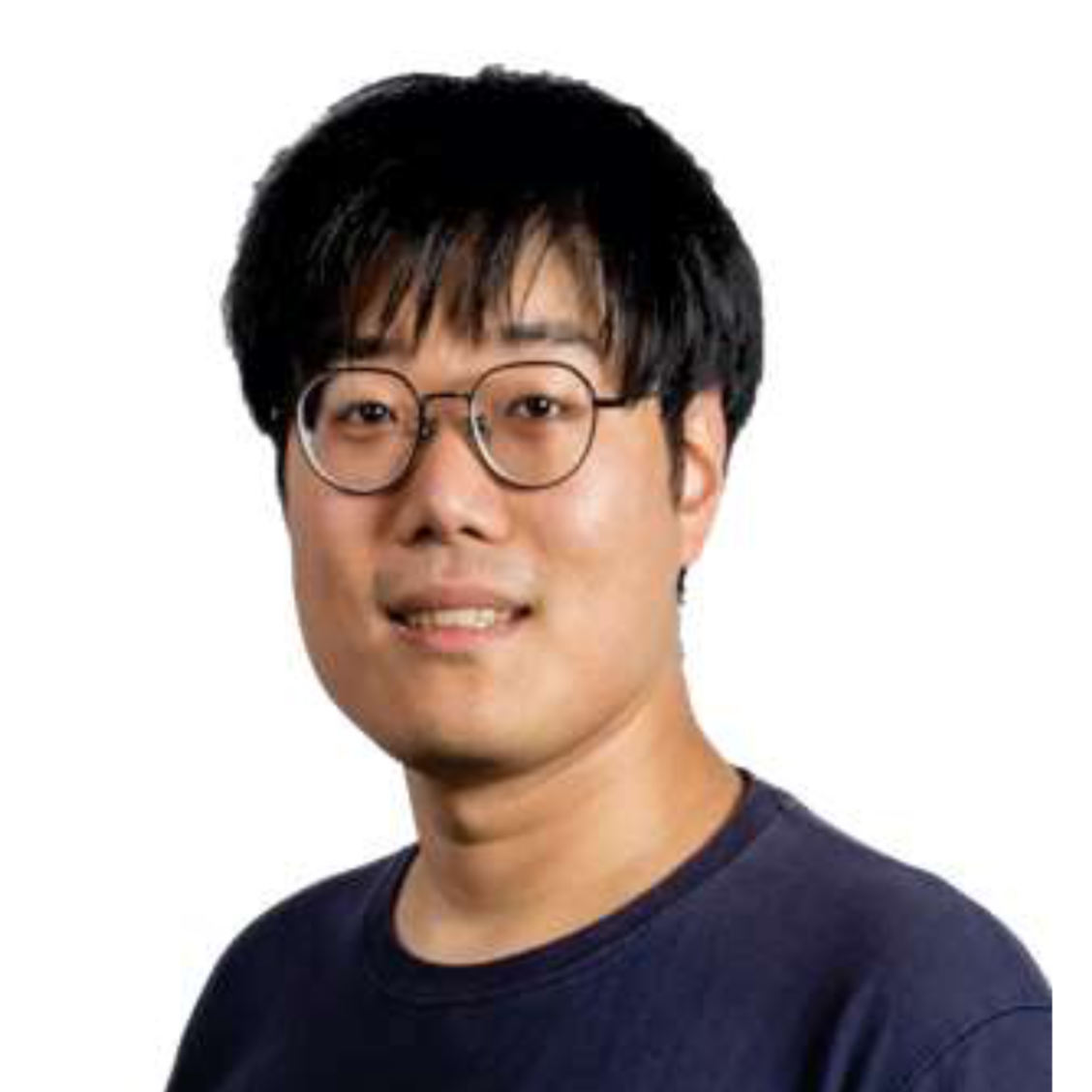}}]{Sanghyuk Chun} is a research scientist at NAVER AI Lab. He was a research engineer at advanced recommendation team (ART) in Kakao Corp from 2016 to 2018. He received his Master's and Bachelor's degree in electronical engineering from KAIST in 2016, 2014, respectively. His research interests include reliable and multi-modal machine learning.
\end{IEEEbiography}

\begin{IEEEbiography}[{\includegraphics[width=1in,height=1.25in,clip,keepaspectratio]{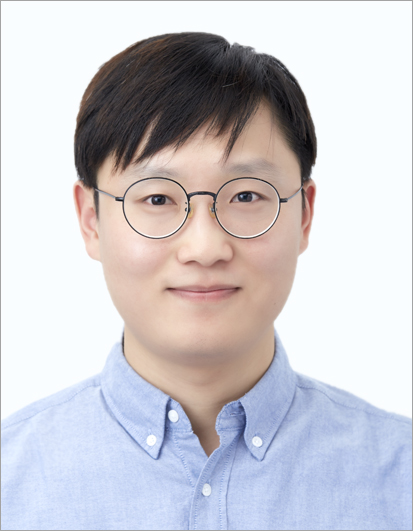}}]{Seungho Lee} is currently a Ph.D. student at the School of Integrated Technology, Yonsei University, Seoul, Korea. He received his B.S. degree in computer science from Yonsei University. His research interests are in the areas of computer vision and deep learning, especially semantic segmentation, image classification, weakly supervised learning, and image understanding.
\end{IEEEbiography}

\begin{IEEEbiography}[{\includegraphics[width=1in,height=1.15in,clip,keepaspectratio]{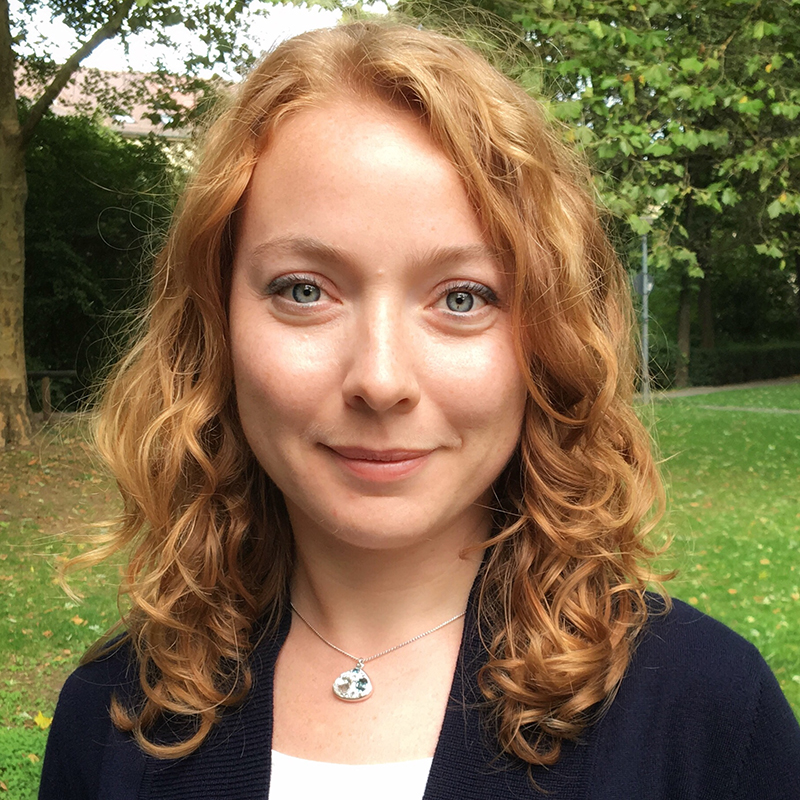}}]{Zeynep Akata} is a professor of Computer Science within the Cluster of Excellence Machine Learning at the University of Tübingen. After completing her PhD at INRIA Rhone Alpes (2014), she worked as a post-doctoral researcher at the Max Planck Institute for Informatics (2014-17) and at University of California Berkeley (2016-17). Before moving to Tübingen in October 2019, she was an assistant professor at the Amsterdam Machine Learning lab of the University of Amsterdam (2017-19). She received a Lise-Meitner Award for Excellent Women in Computer Science from Max Planck Society in 2014, a young scientist honour from the Werner-von-Siemens-Ring foundation in 2019 and an ERC-2019 Starting Grant from the European Commission. Her research interests include multimodal and explainable machine learning.
\end{IEEEbiography}

\begin{IEEEbiography}[{\includegraphics[width=1in,height=1.25in,clip,keepaspectratio]{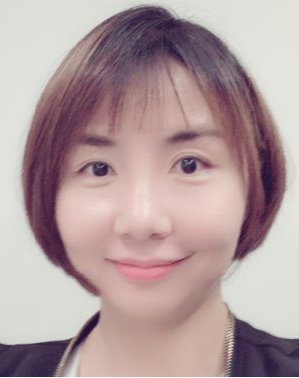}}]{Hyunjung Shim} received her B.S. degree in electrical engineering from Yonsei University, Seoul, Korea, in 2002, and her M.S. and Ph.D. degrees in electrical and computer engineering from Carnegie Mellon University, Pittsburgh, PA, USA, in 2004 and 2008, respectively. 
She was with Samsung Advanced Institute of Technology, Samsung Electronics Company, Ltd., Suwon, Korea, from 2008 to 2013. She is currently an associate Professor with the School of Integrated Technology, Yonsei University. Her research interests include generative models, deep neural networks, classification/recognition algorithms, 3-D vision, inverse rendering, face modeling, and medical image analysis.
\end{IEEEbiography}

\end{document}